\documentclass[twocolumn]{article}

\usepackage{PRIMEarxiv}

\usepackage[utf8]{inputenc} 
\usepackage[T1]{fontenc}    
\usepackage{hyperref}       
\usepackage{url}            
\usepackage{booktabs}       
\usepackage{amsfonts}       
\usepackage{nicefrac}       
\usepackage{microtype}      
\usepackage{lipsum}
\usepackage{fancyhdr}       
\usepackage{graphicx}       
\graphicspath{{media/}}     

\usepackage{amsmath,amssymb,amsfonts}
\usepackage{algorithmic}
\usepackage{graphicx}
\usepackage{textcomp}
\usepackage{mathptmx} 
\usepackage{multirow}
\usepackage{enumerate}
\usepackage{color,soul}
\sethlcolor{yellow}

\usepackage{times}
\usepackage{subcaption}

\usepackage[numbers]{natbib}
\usepackage{booktabs}
\usepackage{lscape}
\usepackage{comment}
\usepackage{verbatim}
\usepackage{ltablex}

\usepackage{stfloats}
\usepackage{lscape}

\usepackage{wrapfig}

\usepackage{longtable}

\usepackage{varwidth}
\usepackage{xcolor, colortbl}

\usepackage{enumitem}

\title{Statistical and Machine Learning Models for Predicting Fire and Other Emergency Events}

\author{
  Dilli Prasad Sharma, Nasim Beigi-Mohammadi, Hongxiang Geng, Alberto Leon-Garcia  \\
  The Edward S. Rogers Sr. Department of Electrical and Computer Engineering \\
  University of Toronto, 10 King's College Road, Toronto, ON M5S 3G4, Canada.\\
  \texttt{\{dilli.sharma, nasim.beigi\}@utoronto.ca}  \\ 
  \texttt{ kevin.geng@mail.utoronto.ca, alberto.leongarcia@utoronto.ca} \\
   \And
  Dawn Dixon, Rob Madro \\
  Fire Rescue Services, City of Edmonton, Alberta, Canada \\
  \texttt{\{dawn.dixon, rob.madro\}@edmonton.ca} \\
   \AND
 Phil Emmenegger, Carlos Tobar, Jeff Li \\
  TELUS Communications Inc, Canada \\
  \texttt{\{phil.emmenegger, carlos.tobar, jeff.li\}@telus.com} 
}

\begin{document}
\twocolumn[ {%
  \begin{@twocolumnfalse}
   \maketitle
   \begin{abstract}
Emergency events in a city cause considerable economic loss to individuals, their families, and the community. Accurate and timely prediction of events can help the emergency fire and rescue services in preparing for and mitigating the consequences of emergency events. In this paper, we present a systematic development of predictive models for various types of emergency events in the City of Edmonton, Canada. We present methods for (i) data collection and dataset development; (ii) descriptive analysis of each event type and its characteristics at different spatiotemporal levels; (iii) feature analysis and selection based on correlation coefficient analysis and feature importance analysis; and (iv) development of prediction models for the likelihood of occurrence of each event type at different temporal and spatial resolutions.  We analyze the association of event types with socioeconomic and demographic data at the neighborhood level, identify a set of predictors for each event type, and develop predictive models with negative binomial regression. We conduct evaluations at neighborhood and fire station service area levels. Our results show that the models perform well for most of the event types with acceptable prediction errors for weekly and monthly periods. The evaluation shows that the prediction accuracy is consistent at the level of the fire station, so the predictions can be used in management by fire rescue service departments for planning resource allocation for these time periods. We also examine the impact of the COVID-19 pandemic on the occurrence of events and on the accuracy of event predictor models.  Our findings show that COVID-19 had a significant impact on the performance of the event prediction models. 
\end{abstract}

\end{@twocolumnfalse} 
}
]
\footnotetext[0]{This work has been submitted to the IEEE for possible publication. Copyright may be transferred without notice, after which this version may no longer be accessible.}

\section{Introduction} \label{label:intro}
Emergency response management is a crucial problem faced by fire and emergency departments around the world. First responders must respond to various events such as fires, rescue, traffic accidents, and medical emergencies.  In recent years, researchers have developed statistical, analytical, and artificial intelligence, specifically, machine learning approaches for designing emergency response management (ERM) systems ~\cite{Mukhopadhyay2022,mukhopadhyay2023artificial}. Emergency incidents in urban areas are often caused by natural disasters (e.g., flooding, earthquake, tsunami), accidents (e.g., fire, explosion, traffic), or pandemic situations. There is a huge socioeconomic impact of fire and emergency incidents due to the considerable economic losses to individuals, families, and communities. Machine learning (ML) models with data obtained from multiple sources such as census bureau, police department, historical fire and rescue events, property inspections and violations, property assessments, property characteristics, parcel, tax lien, and community survey data can help predict the fire risk of the commercial buildings in a city \cite{Madaio2016, SinghWalia2018}. 

A good emergency event predictive model needs to capture spatial and temporal correlations. Neural network-based models such as Convolutions Neural Networks (CNNs), and Recurrent Neural Networks (RNNs) can represent spatio-temporal relationships for the event incident locations and rates \cite{ren2018deep, yuan2018hetero}. However, these models are limited to specific events (e.g., structural fire, traffic accidents) and lose their accuracy for small spatial regions of a city and short time intervals. Predicting the events occurring in neighborhoods, wards, or suburbs of the city for a certain time duration (e.g., daily, weekly, or monthly) would help with emergency planning and risk mitigation.  The characteristics of the city such as population density, land and property/building structure types and occupancy, household income, length of residency, education, age group, and point of interest can influence the likelihood of occurrence of the events. Identifying the key features/variables and assessing their association with the events can be useful for predicting the risk of event occurrence in the city.  

In this paper, we present our approach using statistical analysis and machine learning to predict the likelihood of fire and other emergency events in the City of Edmonton, Alberta. We strive to answer the following research questions:
\begin{enumerate}
    \item  What data and sources are relevant to fire and emergency event occurrence?
    \item  How can we engineer features from various data sources across various temporal and spatial granularities?
    \item  What features are highly correlated with fire and other emergency events?
    \item  Can we develop a model that predicts fire and other emergency events with good accuracy given the features?
    \item  Can the prediction model predict all types of events? or can it predict some better than others?
    \item  Can the model predict events at any level of spatial and temporal granularity? For instance, can the model be used across all fire stations?
    \item Can we predict the events at any temporal level (i.e., hourly, daily, monthly, etc.)?
    \item How does the COVID-19 pandemic impact the prediction of emergency events?
    
\end{enumerate}

To address the above questions,  we present a systematic methodology for (i) dataset development and feature extraction, (ii) feature analysis and selection, and (iii) model development and evaluations for predicting the Edmonton Fire Rescue Service (EFRS) events at neighborhood and fire-station levels for the City of Edmonton. We collected the city's data from multiple sources, extracted relevant features by applying computational methods to prepare the datasets, and analyzed them with state-of-the-art statistical and machine-learning models. We collected neighborhood demographic and socio-economic information, and EFRS event data from the city's open data portal~\cite{OpenDataPortal_Edmonton,event_data}, and map data (points of interest) from OpenStreetMap\cite{OpenStreetMap}). We made the following {\bf key contributions} in this work:
\begin{itemize}
    \item Development of a systematic approach to collect data from different sources (city, maps, fire, and emergency events, etc.) and preparation of datasets at different temporal and spatial levels for training and validating ML models designed to predict fire and other emergency events.
    
    \item Identifying the key predictor variables that are strongly associated with EFRS event types; we identified highly correlated features for each of the EFRS event types and used them for building the ML models.
    
    \item A descriptive analysis of EFRS event occurrence at different time intervals (i.e., hourly, daily, weekly, monthly, and yearly) at city and neighborhood levels for use in model selection, explanation, and validation of the prediction results.
    
    \item Predictive models for each EFRS event type with the identified key variables. These models predict the likelihood of occurrence of EFRS events by neighborhood and/or fire station. Evaluation of the models using standard regression metrics and prediction error analysis. These predictions can be used for emergency management and planning in neighborhoods and/or fire-station zones.
    
    \item We use weekly and monthly prediction values to classify the neighborhoods according to event risk; this may be useful in decision-making and planning to reduce/prevent the damage associated with events.

    \item We assessed the impact of the COVID-19 pandemic on the performance of EFRS event predictive models using the mean absolute error (MAE) and root mean square error (RMSE) metrics. We compared models trained with before and after COVID-19 event data. The analysis of the impact of the COVID-19 pandemic provides insights on fire and emergency event prediction that can help fire and rescue service departments in their preparedness planning, response, and recovery.

\end{itemize}
The rest of the paper is organized as follows: Section \ref{sec-related-work} presents related work. Section \ref{section-datacollection} discusses data collection and feature extraction. Section \ref{label:efrs-eventanalysis} analyzes the EFRS events. Section \ref{sec-feature-analysis} describes feature analysis and selection.  Section \ref{exerimental-setup} presents the predictive model development. Section \ref{prediction-by-neighborhoods} discusses the analysis of prediction results by neighborhood, and Section \ref{sec-predictions-by-fire-station} by fire-station service area. Section \ref{sec-covidimpact} discusses the impact of the COVID-19 pandemic on EFRS events predictions. Section \ref{sec:insights} presents insights and key findings. Lastly, Section \ref{section-conclusion} provides conclusions and suggests future research directions.
\begin{figure*} [hbt!]
  \centering
    \includegraphics[width=0.95\textwidth,keepaspectratio]{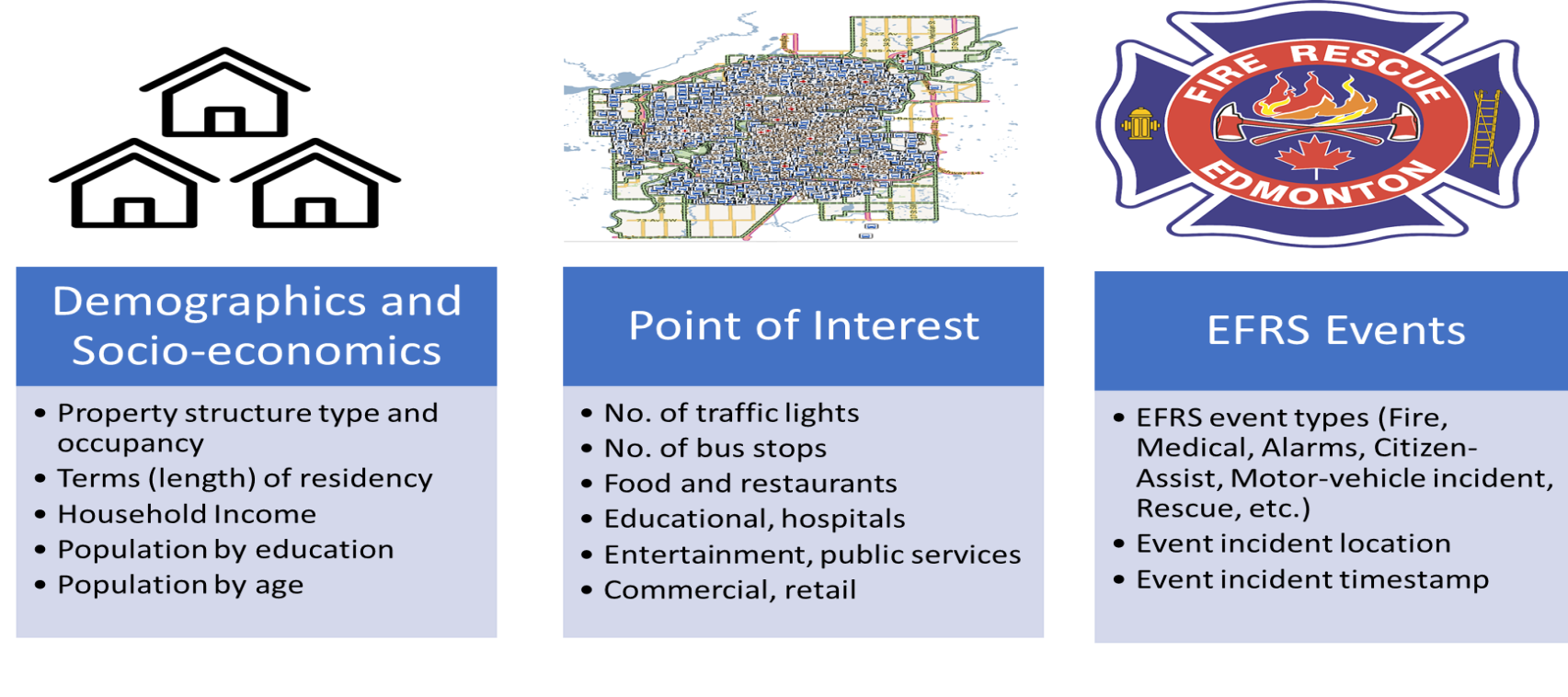}
    \caption{\small Data categories and their sources }
 \label{fig_datasources}
\end{figure*}

\section{Related Work} \label{sec-related-work}

Models to predict and forecast the occurrence of emergency events can have a significant impact in saving people's lives and reducing property damage.  In recent years, researchers have made a number and variety of contributions towards the development of prediction techniques using statistical and artificial intelligence technologies~\cite{Mukhopadhyay2022, Huang2021}.

\citet{Madaio2016} developed a fire risk assessment framework to identify and prioritize commercial property fire inspections using machine learning models. Data collected from multiple sources (Fire Rescue Department, City of Atlanta, Police Department, U. S. Census Bureau) were used to train models. Fire risk scores were computed for buildings in the city of Atlanta, based on predictive models for fire events. Several models were considered: Logistic Regression, Gradient Boosting, Support Vector Machine (SVM), and Random Forest. 

In a similar study, \citet{SinghWalia2018} proposed two fire risk predictive models for commercial and residential properties in Pittsburgh. The models were developed for structural fire risk prediction using historical fire-related incidents, property inspections and violations, property assessments, property characteristics and parcel data, tax lien, and American Community Survey data. The commercial risk model was used to generate a risk score (i.e., likelihood of fire) for non-residential (commercial, governmental, industrial, etc.) properties in the city. The commercial risk model was deployed at the Pittsburgh Bureau of Fire. The residential risk model was used to predict and generate fire risk scores for residential census blocks in Pittsburgh. However, both of these studies~\cite{Madaio2016,SinghWalia2018} mainly focused on fire events only.

\citet{Ye2022} proposed a Finite Element (FE)-based machine learning (ML) framework to predict structural fire response that integrates the advantages of accuracy from FE methods as well as efficiency from ML techniques. Numerical data of a steel frame structure with temperature data were used to train the ML models. Four ML models including Support Vector Machine (SVM), Decision Tree (DT), Random Forest (RF), and Gradient Boosting (GB) were used in this study. However, the scope of this study is limited to the structural fire of a building. 

\citet{ren2018deep}, and \citet{yuan2018hetero} proposed Convolutional Neural Networks (CNNs) and Recurrent Neural Networks (RNNs)-based models to predict traffic accidents, where CNNs and RNNs models are used to capture the spatial and temporal correlations, respectively. The models predict the occurrence of traffic accidents at a city level. However, these models lose their accuracy for short time intervals and small study zones. One of the important causes of this performance degradation is the zero-inflated problem (i.e., rare occurrence of accidents), which becomes more dominant as the spatiotemporal resolution increases \cite{bao2019spatiotemporal}. Recently, \citet{zhou2020riskoracle} developed a promising set of solutions for tackling the zero-inflated problem with some useful guidelines for improving the performance of accident predictions in high temporal resolution. However, the emergency events considered were limited to traffic accidents. 

Data-driven predictive models are widely used in many prior works\cite{Agarwal2020,Cortez2018,Roque2022}. \citet{Agarwal2020} studied the impact of weather on the damage caused by fire incidents in the United States. The authors prepared a dataset by collecting fire incidents and weather data from sources such as the National Fire Incident Reporting System and the National Oceanic and Atmospheric Administration, respectively. They then developed predictive models to analyze the fire risk. A Gradient Boosting Tree (GBT) model was used to predict the losses due to fire incidents. The paper presents useful insights to fire and rescue managers and researchers with a detailed framework of big data and predictive analytics for effective management of fire risk. However, it is limited to the analysis of weather data and their impact on fire events.

\citet{Cortez2018} proposed a Recurrent Neural Networks (RNN)-based prediction model with a Long Short-Term Memory (LSTM) architecture for the prediction of emergency events. Spatial clustering, spatial analysis, and LSTM-based prediction techniques were integrated to develop this model. The model was compared to autoregressive integrated moving average (ARIMA) models using the historical emergency event data provided by the national police of Guatemala. Similarly, \citet{Roque2022} developed a neural network-based emergency event prediction model for the city of Florianópolis, Brazil. Spatial clustering and Long-Short Term Memory (LSTM) methods were combined to develop their predictive model. However, the predictions were limited to two types of observations (pre-hospital care and others) in four regions of the city (4 clusters). Finally, \citet{Jin2020}  proposed a deep sequence learning-based fire situation forecasting network to predict the incidents. The network integrates the structures of Variational auto-encoders and sequence generative models to obtain the latent representation of the fire situation and to learn spatiotemporal data from GIS to train a generative model. The study used fire incident data from the city of San Francisco.

The state-of-the-art research on fire and emergency event prediction has mainly focused on structural fire and/or medical events at high levels of spatial granularity (e.g., city level or large regions), and the models that have been developed targetted a specific consideration, such as commercial areas or buildings, or a certain event type such as structure fire, medical, etc. In addition, the models lack the systematic methodology for data collection and feature extraction, feature analysis and selection, and model development and evaluation.  In this paper, we present a systematic methodology that describes the life-cycle of statistical analysis and machine learning development to predict fire and other emergency events not only in large geographical regions such as cities but also at neighborhood and fire station service area levels. 

We investigate the association of different data sources with fire and other emergency events and develop a systematic approach to collect the data from a variety of sources. The predictive models are developed for the prediction of weekly and monthly event incidents at a small area of a city (fire-station regions, neighborhood levels) as well. 

\begin{figure*} [hbt!]
  \centering
    \includegraphics[width=0.95\textwidth,keepaspectratio]{ 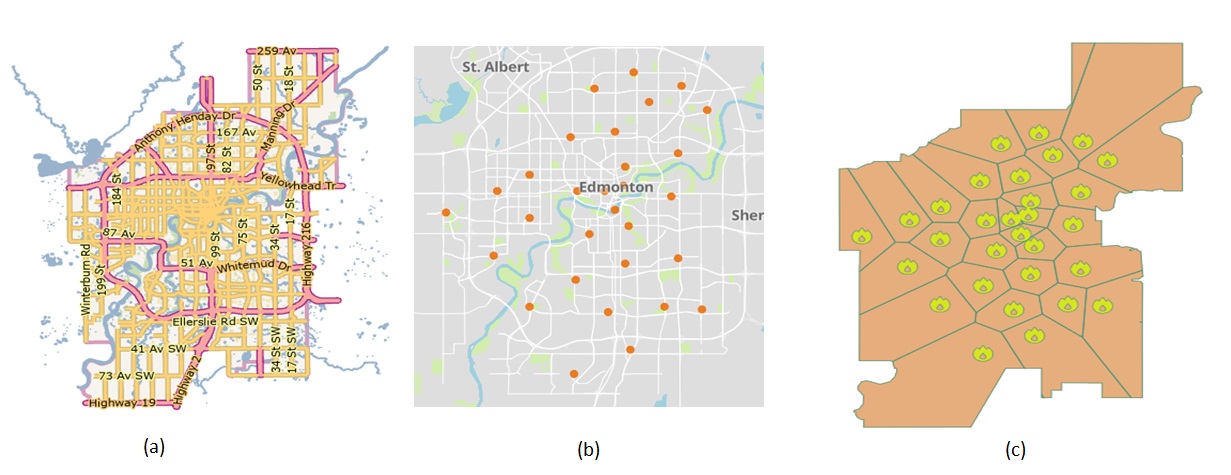}
    \caption{\small (a) Neighborhoods, (b) Fire-station locations, and (c) Fire-station Voronoi regions }
 \label{fig_neighborhood-fs}
\end{figure*}

\section{Data Collection and Feature Extraction} \label{section-datacollection}

 The City of Edmonton is the capital of the Alberta province of Canada with a population of 1,010,899 in 2021 \cite{population_in_2021}.  We collected data for the city from: the City of Edmonton Open Data Portal ~\cite{OpenDataPortal_Edmonton}, and 
 OpenStreetMap \cite{OpenStreetMap}. From these two data sources, we created three categories of data: 1. From the city's open data portal we collected neighborhood demographic and socioeconomic data; 2. We used OpenStreetMap to collect neighborhood Point of Interest and road related features; and 3. we also used the city's open data portal to collect Edmonton Fire Rescue Service Event data. We now discuss the details of these data.
 

\subsection{Data Categories} \label{data-sources}
Data collection and feature extraction is an important part of developing predictive models. Since various factors can be involved in causing emergency events, the main step in building an event prediction model is to collect as much relevant data as possible from available sources~\cite{Mukhopadhyay2022}. We extracted features from the following three categories of data.

\begin{itemize}
\item {\bf City of Edmonton Open Data Portal}: We collected neighborhood demographic and socioeconomic data from the Open Data Portal of the City of Edmonton~\cite{OpenDataPortal_Edmonton}. The portal provides diverse demographic information of the neighborhoods such as building/property type and occupancy~\cite{population_by_unit_structure_types,population_by_occupancy}, length of residency ~\cite{population_length_of_residency}, household income~\cite{population_by_householdincome}, population by education~\cite{population_by_education}, and population by age~\cite{population_by_age}. All demographic information by neighborhood is based on the Census 2016. The descriptions of the selected features are as follows: 
\begin{itemize}[nosep]
\setlength\itemsep{0em}
    \item {\bf Apartment/Condo (1-4 Stories)}: number of units occupied in apartments or condos with 1 to 4 stories.
    \item {\bf Apartment/Condo (5+ Stories)}: number of units occupied in apartments or condos with 5 or more stories.
    \item {\bf Duplex/Fourplex}: number of units occupied in duplex or fourplex houses. 
    \item {\bf Hotel/Motel}: number of units occupied in hotels or motels.
    \item {\bf Institution/Collective Residence}: number of units occupied in institution and collective residences.
    \item {\bf Manufactured/Mobile Home}: number of units occupied in manufactured or mobile homes.
    \item {\bf RV/Tent/Other}: number of units occupied in RVs, tents, or other houses.
    \item {\bf Row House}: number of units occupied in row houses.
    \item {\bf Single Detached House}: number of units occupied in single detached houses.
    \item {\bf 5 Years or More}: number of respondents who have been staying for 5 or more years.  
    \item {\bf 3 Years to Less than 5 Years}: number of respondents who have been staying 3 years to less than 5 years.  
    \item {\bf 1 Year to Less than 3 Years}: number of respondents who have been staying more than 1 year but less than 3 years.  
    \item {\bf Less than 1 Year}: number of respondents who have been staying less than 1 year.
    \item {\bf Low Income}: number of people in households with less than $30k$ income.
    \item {\bf Low-medium Income}: number of people in households whose income ranges from $30k$ to $100k$.
    \item {\bf Medium Income}: number of people in households with income ranges from $100k$ to $200k$.
    \item {\bf High Income}: number of people in households with income greater than $200k$.
    \item {\bf No Certificate, Diploma or Degree}: number of people with no certificate, diploma or degree attained a level of education.
    \item {\bf High School, Trades or Apprenticeship Certificate:} number of people for high school, trades, or apprenticeship certificates attained the level of education.
    \item {\bf College or University Certificate or Diploma}: number of persons with college or university certificate or diploma attained level of education
    \item {\bf University Bachelor and Medical Degree}: number of people with university bachelor's or medical degrees attained level of education.
    \item {\bf Master and Doctorate Degree}: number of people with a master or doctoral degree attained level of education.
    \item {\bf Children}: number of people of age less than $15$ years.
    \item {\bf Youth}: number of people in age group from $15$ to $24$ years.
    \item {\bf Adults}: number of people in age group from $25$ to $64$ years.
    \item {\bf Seniors}: number of people in age $65$ or above.
    \item {\bf Population}: population per square kilometer.
\end{itemize}

\begin{figure*} [h!]
		\centering
		\includegraphics[width=0.6\textwidth, height=0.45\textwidth]{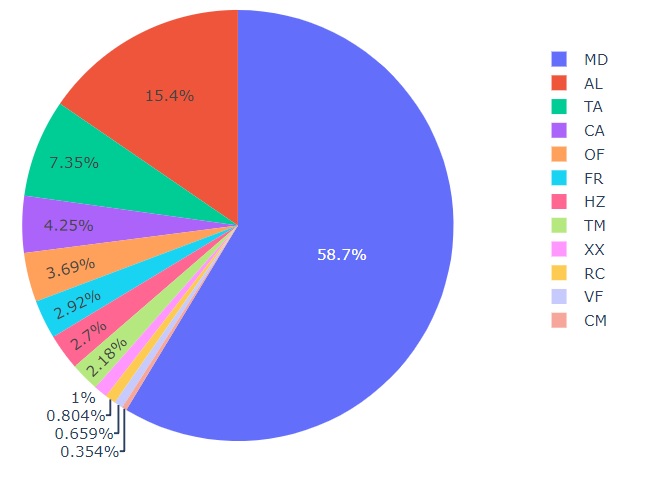}
        \caption{Occurrence of the Different Event Types}
        \label{events_occurance_percentages}
\end{figure*}

\item {\bf Open Street Map (OSM)}: We collected Point of Interest (POI) and road-related features from the OpenStreetMap~\cite{OpenStreetMap} for amenity, building and highway categories using the Overpass API~\cite{OSMOverpassQL}. We extracted the number of POI facilities (e.g., pubs, restaurants, schools/colleges, etc.), and road-related features for different sub/categories of amenities, buildings, and high-way types for each neighborhood of the City of Edmonton. We entered a query for each category to the Overpass API~\cite{OSMOverpassQL}. We grouped POI subcategories into a single feature based on their similarities. For example, pubs, bars, restaurants, food courts, etc. are food-related amenities. Nine different features were collected from the OSM sources which are as follows:
\begin{itemize}
    \item {\bf Food}: number of food-related facilities including bars, cafes, fast foods, food courts, pubs, and restaurants.
    \item {\bf Education}: number of educational institutions including college, kindergarten, library, school, and university.
    \item {\bf Healthcare}: number of healthcare facilities such as clinics and hospitals.
    \item{\bf Entertainment}: number of entertainment-related facilities including art centers, cinemas, community centers, events venues, nightclubs, and theatres.
    \item {\bf Public Service}: number of public service-related items including the courthouse, fire stations, police, and town halls.
    \item {\bf Commercial}: number of commercial buildings including offices, commercial, and government.
    \item {\bf Retail}: number of retail buildings.
    \item {\bf No. traffic\_lights}: number of traffic lights. 
    \item {\bf No. bus\_stops}: number of bus stops. 
\end{itemize}

\item {\bf Edmonton Fire Rescue Service (EFRS) Events}: We collected EFRS event data from the Open Data Portal of the City of Edmonton~\cite{OpenDataPortal_Edmonton,event_data}. This data consists of information for current and historical EFRS events including event type (category/subcategory), event dispatch time, location, response code, and equipment. The EFRS event category/type includes fire events, medical events, alarm events, citizen assist events, motor vehicle incident events, rescue events, outside fire events, vehicle fire events, community events, training/maintenance events, hazardous material events, and other events.
\end{itemize}

\subsection{Dataset Preparation} \label{sec-data-preparation}

We collected data based on the current 402 city neighborhoods including both residential and industrial areas in the City of Edmonton. Each neighborhood's polygon boundary is shown in Figure~\ref{fig_neighborhood-fs} (a) where the yellow line indicates the neighborhood's boundary. The static features (socioeconomic and demographic data) for each neighborhood are collected first.  Then the neighborhood features are mapped to the service region for each fire station. There are 30 fire stations in the city and the locations are shown in Figure~\ref{fig_neighborhood-fs} (b). We approximate the service region for each fire station by finding the Voronoi diagram that maps each point in the city to the nearest fire station~\cite{Aurenhammer1991}. In general, each such service area includes several neighborhoods. Figure~\ref{fig_neighborhood-fs} (c) shows the Voronoi region for each fire station. A neighborhood can straddle multiple fire-station regions. We computed the overlap proportion for the neighborhood that has straddled multiple fire stations and distributed its feature values to its associated fire stations. We merged all demographic, socioeconomic, PoIs, and EFRS events data for different time intervals to obtain feature vectors for each neighborhood and for each fire station.
\section{Descriptive Analysis of EFRS Events}
\label{label:efrs-eventanalysis}
In this section, we present a descriptive analysis of event occurrence properties.  We first describe the different EFRS event types and the volume of associated data samples. Next we present the descriptive statistics for different event types at the city and neighborhood levels over different time intervals: hourly, daily, weekly, monthly, and yearly. For this analysis, we considered event data for the $5$ years from $2016$ to $2021$.

\subsection{EFRS Event Types } \label{Event-types}
The EFRS event data contains information about event incidents in the City of Edmonton including: event dispatched time-stamp, event close-time, location, event category types, etc. for the period 2011-01-01 to 2022-08-30 ~\cite{OpenDataPortal_Edmonton,event_data}. The EFRS event types are as follows:
\begin{itemize}[nosep]
    \setlength\itemsep{0em}
    \item Medical (MD)
    \item Alarms (AL) 
    \item Motor vehicle incident (TA)
    \item Citizen assist (CA)
    \item Outside fire (OF)
    \item Fire (FR) 
    \item Hazardous materials (HZ)
    \item Training/maintenance (TM)
    \item Rescue (RC)
    \item Vehicle fire (VF)
    \item Community (CM)
    \item Others (XX)
\end{itemize}

Figure~\ref{events_occurance_percentages} shows the proportion of the total event occurrences for different types. Medical events are by far the most frequent events occupying 58.7\% of the event incidents. Alarms (15.4\%) and Traffic incidents (7.35\%) events are the second and third most frequent events, respectively. Fire events are relatively less frequent (6th) occurring at 2.92\%.  
\begin{figure*}  
\centering
\subfloat[Fire (left) and Medical (right)] {
 \includegraphics[width=0.95\textwidth]{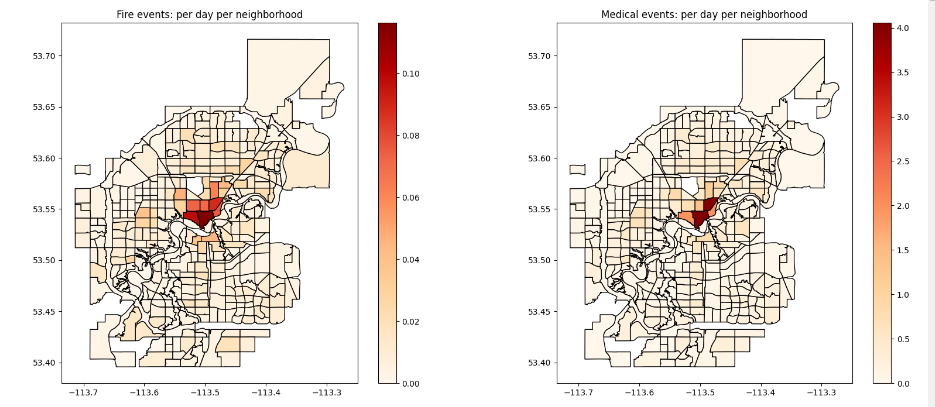}} 

\subfloat[Alarms (left) and Citizen Assist (right)]
{
 \includegraphics[width=0.95\textwidth]{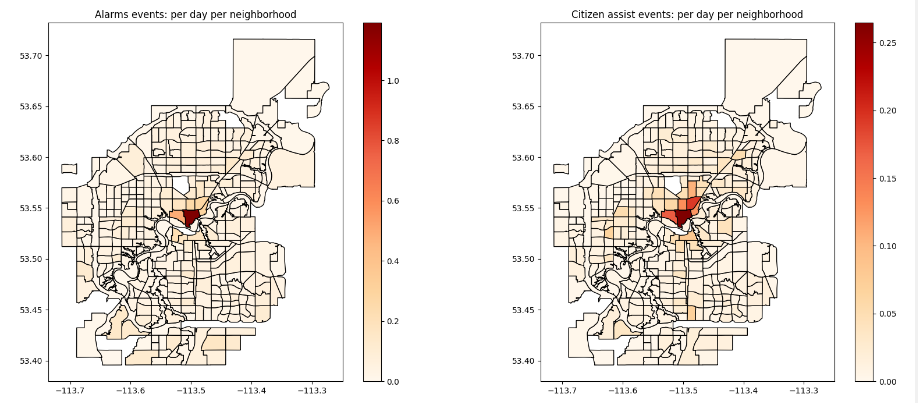}}

 \subfloat[Motor-Vehicle Incidents (left) and Rescue]
 {
 \includegraphics[width=.95\textwidth]{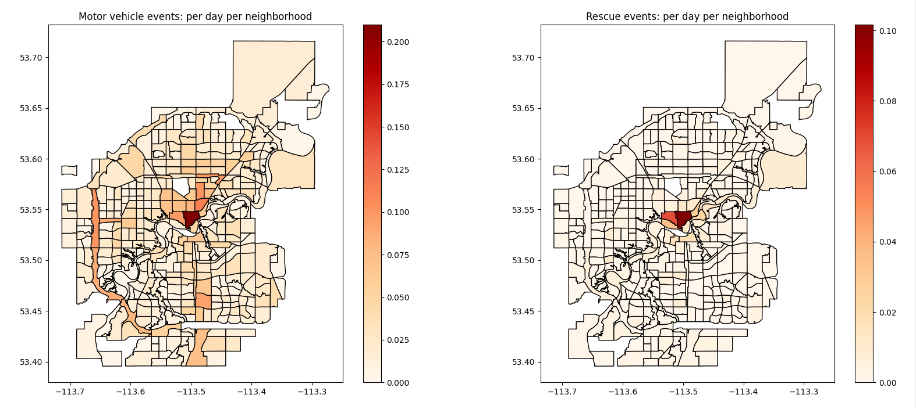}
}

\caption{Plot of daily event occurrence rate by neighborhood}
\label{fig-events-in-neighborhood}
\end{figure*}

\subsection{Descriptive Analysis of Events} \label{discriptive-analysis}

In this section, we present a descriptive statistical analysis of the six event types. The analysis helps to identify the event occurrence patterns in terms of statistical metric such as the mean, standard deviation, and coefficient of variation of the occurrence of event types in different time periods. The coefficient of variation (CV) is the standard deviation over the mean and is a measure of the dispersion of data points about the mean and is useful for comparing the degree of variability of different event types. 

Table ~\ref{tab-descriptive-statistics of events} presents the descriptive statistics for six event types over the different periods of time intervals in the city. These statistics are computed with $5$ years of event data from $2016$ to $2021$. A fire event incident happens every $7$ hours on average; alarm events happen every $1$ hour; citizen assist events every $4$ hours; traffic incident events every $2$ hours; rescue events every $20$ hours, and $3.66 ~(\approx 4) $ medical events happen every hour. The CV of the event occurrence shows that there is a high degree of variability in the number event occurrences in shorter time periods. In contrast the CV approaches zero over longer time intervals indicating small variability about the mean.  

Medical events are relatively consistent with CV less than 1 for all time intervals. Fire events have a significantly larger CV than most other events, suggesting that the prediction of number of fire events for short time intervals will be difficult. Rescue events have even more variation than fire events.  We note that the mean number of event occurrences is highest for medical, with fire next, and then rescue. 

\begin{table}[]
\centering
\captionsetup{justification=centering}
\caption{ \small Descriptive statistics for the EFRS events at a city level.} 
\label{tab-descriptive-statistics of events}
\resizebox{0.95\columnwidth}{!}{%
\begin{tabular}{|
>{\columncolor[HTML]{FFFFFF}}c |
>{\columncolor[HTML]{FFFFFF}}l |
>{\columncolor[HTML]{FFFFFF}}l |
>{\columncolor[HTML]{FFFFFF}}l |
>{\columncolor[HTML]{FFFFFF}}l |}
\hline
\cellcolor[HTML]{EFEFEF}Event Type &
  \cellcolor[HTML]{EFEFEF}Time Interval &
  \cellcolor[HTML]{EFEFEF}Mean ($\mu$) &
  \cellcolor[HTML]{EFEFEF}{\color[HTML]{343434} StdDev ($\sigma$)} &
  \cellcolor[HTML]{EFEFEF} CV ($\sigma/\mu$) \\ \hline
\cellcolor[HTML]{FFFFFF} &

  \cellcolor[HTML]{FFFFFF}Hourly &
  \cellcolor[HTML]{FFFFFF}0.15 &
  {\color[HTML]{343434} 0.40} &
  \cellcolor[HTML]{FFFFFF} 2.55 \\ \cline{2-5} 
\cellcolor[HTML]{FFFFFF} &
  \cellcolor[HTML]{FFFFFF}Daily &
  \cellcolor[HTML]{FFFFFF}3.81 &
  {\color[HTML]{343434} 2.08} &
  \cellcolor[HTML]{FFFFFF}0.55\\ \cline{2-5} 
\cellcolor[HTML]{FFFFFF} &
  \cellcolor[HTML]{FFFFFF}Weekly &
  \cellcolor[HTML]{FFFFFF}26.62 &
  {\color[HTML]{343434} 6.42} &
  \cellcolor[HTML]{FFFFFF}0.24 \\ \cline{2-5} 
\cellcolor[HTML]{FFFFFF} &
  \cellcolor[HTML]{FFFFFF}Monthly &
  \cellcolor[HTML]{FFFFFF}116.26 &
  {\color[HTML]{343434} 15.34} &
  \cellcolor[HTML]{FFFFFF}0.13 \\ \cline{2-5} 
\multirow{-5}{*}{\cellcolor[HTML]{FFFFFF}Fire} &
  \cellcolor[HTML]{FFFFFF}Yearly &
  \cellcolor[HTML]{FFFFFF}1395.20 &
  {\color[HTML]{343434} 94.89} &
  \cellcolor[HTML]{FFFFFF}0.07 \\ \hline

\cellcolor[HTML]{FFFFFF} &
  Hourly &
  3.66 &
  {\color[HTML]{343434} 2.18} &
  0.60 \\ \cline{2-5} 
\cellcolor[HTML]{FFFFFF} &
  Daily &
  87.90 &
  {\color[HTML]{343434} 13.28} &
  0.15\\ \cline{2-5} 
\cellcolor[HTML]{FFFFFF} &
  Weekly &
  612.98 &
  {\color[HTML]{343434} 63.41} &
  0.10 \\ \cline{2-5} 
\cellcolor[HTML]{FFFFFF} &
  Monthly &
  2676.70 &
  {\color[HTML]{343434} 214.27} &
  0.08 \\ \cline{2-5} 
\multirow{-5}{*}{\cellcolor[HTML]{FFFFFF}Medical} &
  Yearly &
  32120.40 &
  {\color[HTML]{343434} 2074.58} &
  0.06 \\ \hline

\cellcolor[HTML]{FFFFFF} &
  Hourly &
  0.92 &
  {\color[HTML]{343434} 1.11} &
  1.20 \\ \cline{2-5} 
\cellcolor[HTML]{FFFFFF} &
  Daily &
  22.14 &
  {\color[HTML]{343434} 7.41} &
  0.33\\ \cline{2-5} 
\cellcolor[HTML]{FFFFFF} &
  Weekly &
  154.43 &
  {\color[HTML]{343434} 35.71} &
  0.23\\ \cline{2-5} 
\cellcolor[HTML]{FFFFFF} &
  Monthly &
  674.38 &
  {\color[HTML]{343434} 118.92} &
  0.18\\ \cline{2-5} 
\multirow{-5}{*}{\cellcolor[HTML]{FFFFFF}Alarms} &
  Yearly &
  8092.60 &
  {\color[HTML]{343434} 1106.22} &
  0.14 \\ \hline

\cellcolor[HTML]{FFFFFF} &
  Hourly &
  0.24 &
  {\color[HTML]{343434} 0.53} &
   2.24 \\ \cline{2-5} 
\cellcolor[HTML]{FFFFFF} &
  Daily &
  5.76 &
  {\color[HTML]{343434} 3.42} &
  0.59 \\ \cline{2-5} 
\cellcolor[HTML]{FFFFFF} &
  Weekly &
  40.22 &
  {\color[HTML]{343434} 15.01} &
  0.37 \\ \cline{2-5} 
\cellcolor[HTML]{FFFFFF} &
  Monthly &
  175.66 &
  {\color[HTML]{343434} 53.56} &
  0.30 \\ \cline{2-5} 
\multirow{-5}{*}{\cellcolor[HTML]{FFFFFF}Citizen Assist} &
  Yearly &
  2108.00 &
  {\color[HTML]{343434} 126.72} &
  0.06 \\ \hline

\cellcolor[HTML]{FFFFFF} &
  Hourly &
  0.50 &
  {\color[HTML]{343434} 0.83} &
  1.68 \\ \cline{2-5} 
\cellcolor[HTML]{FFFFFF} &
  Daily &
  11.97 &
  {\color[HTML]{343434} 5.70} &
  0.48 \\ \cline{2-5} 
\cellcolor[HTML]{FFFFFF} &
  Weekly &
  83.47 &
  {\color[HTML]{343434} 24.47} &
  0.29 \\ \cline{2-5} 
\cellcolor[HTML]{FFFFFF} &
  Monthly &
  364.50 &
  {\color[HTML]{343434} 81.18} &
  0.22 \\ \cline{2-5} 
\multirow{-5}{*}{\cellcolor[HTML]{FFFFFF}\begin{tabular}[c]{@{}c@{}}Motor \\ Vehicle \\ Incident\end{tabular}} &
  Yearly &
  4374.00 &
  {\color[HTML]{343434} 773.18} &
  0.18 \\ \hline

\cellcolor[HTML]{FFFFFF} &
  Hourly &
  0.05 &
  {\color[HTML]{343434} 0.22} &
  4.65 \\ \cline{2-5} 
\cellcolor[HTML]{FFFFFF} &
  Daily &
  1.17 &
  {\color[HTML]{343434} 1.12} &
  0.96 \\ \cline{2-5} 
\cellcolor[HTML]{FFFFFF} &
  Weekly &
  8.16 &
  {\color[HTML]{343434} 3.27} &
  0.40 \\ \cline{2-5} 
\cellcolor[HTML]{FFFFFF} &
  Monthly &
  35.66 &
  {\color[HTML]{343434} 9.09} &
  0.26 \\ \cline{2-5} 
\multirow{-5}{*}{\cellcolor[HTML]{FFFFFF}Rescue} &
  Yearly &
  428.00 &
  {\color[HTML]{343434} 25.96} &
  0.06 \\ \hline
\end{tabular}%
}
\end{table}
\subsection{Events Occurrence by Neighborhood } \label{daily-events-by-neighborhoods}
Table \ref{descriptive-statistics of events by neighborhood} shows the descriptive statistics of the rate of occurrence for six different event types by neighborhood with different temporal resolutions. As before the CV decreases as the time intervals increase. What stands out is the the CV values per neighborhood are much larger than citywide. Fire and rescue events have significantly large degrees of variation underscoring the difficulty in predicting the number of events in small geographic areas.  Even medical have significant variability at the neighborhood level.  variations at the neighborhood level. These results show that the prediction of the number of occurrences of individual event types at the neighborhood level is very difficult due to the low rate of event occurrences. 
\begin{table}[]
\centering
 \captionsetup{justification=centering}
\caption{\small Descriptive statistics for the EFRS event rate at neighborhood level}
\label{descriptive-statistics of events by neighborhood}
\resizebox{0.95\columnwidth}{!}{%
\begin{tabular}{|
>{\columncolor[HTML]{FFFFFF}}c |
>{\columncolor[HTML]{FFFFFF}}l |
>{\columncolor[HTML]{FFFFFF}}l |
>{\columncolor[HTML]{FFFFFF}}l |
>{\columncolor[HTML]{FFFFFF}}l |}
\hline
\cellcolor[HTML]{EFEFEF}Event Type &
  \cellcolor[HTML]{EFEFEF}Time Interval &
  \cellcolor[HTML]{EFEFEF}Mean ($\mu$) &
  \cellcolor[HTML]{EFEFEF}{\color[HTML]{343434} StdDev ($\sigma$)} &
  \cellcolor[HTML]{EFEFEF} CV ($\sigma/\mu$) \\ \hline
\cellcolor[HTML]{FFFFFF} &
  \cellcolor[HTML]{FFFFFF}Hourly &
  \cellcolor[HTML]{FFFFFF}0.0004 &
  {\color[HTML]{343434} 0.0204} &
  \cellcolor[HTML]{FFFFFF} 49.2401 \\ \cline{2-5} 
\cellcolor[HTML]{FFFFFF} &
  \cellcolor[HTML]{FFFFFF}Daily &
  \cellcolor[HTML]{FFFFFF}0.0099 &
  {\color[HTML]{343434} 0.1017} &
  \cellcolor[HTML]{FFFFFF}   10.2281\\ \cline{2-5} 
\cellcolor[HTML]{FFFFFF} &
  \cellcolor[HTML]{FFFFFF}Weekly &
  \cellcolor[HTML]{FFFFFF}0.0693 &
  {\color[HTML]{343434} 0.2897} &
  \cellcolor[HTML]{FFFFFF} 4.1784 \\ \cline{2-5} 
\cellcolor[HTML]{FFFFFF} &
  \cellcolor[HTML]{FFFFFF}Monthly &
  \cellcolor[HTML]{FFFFFF}0.3006 &
  {\color[HTML]{343434} 0.7351} &
  \cellcolor[HTML]{FFFFFF}  2.4456  \\ \cline{2-5} 
\multirow{-5}{*}{\cellcolor[HTML]{FFFFFF}Fire} &
  \cellcolor[HTML]{FFFFFF}Yearly &
  \cellcolor[HTML]{FFFFFF}3.5665 &
  {\color[HTML]{343434} 6.0763} &
  \cellcolor[HTML]{FFFFFF}  1.7037 \\ \hline
\cellcolor[HTML]{FFFFFF} &
  Hourly &
  0.0095 &
  {\color[HTML]{343434} 0.0997} &
    10.4423    \\ \cline{2-5} 
\cellcolor[HTML]{FFFFFF} &
  Daily &
  0.2289 &
  {\color[HTML]{343434} 0.6422} &
    2.8054 \\ \cline{2-5} 
\cellcolor[HTML]{FFFFFF} &
  Weekly &
  1.5962 &
  {\color[HTML]{343434} 3.2154} &
  2.0144 \\ \cline{2-5} 
\cellcolor[HTML]{FFFFFF} &
  Monthly &
  6.9201 &
  {\color[HTML]{343434} 13.0662} &
   1.8882   \\ \cline{2-5} 
\multirow{-5}{*}{\cellcolor[HTML]{FFFFFF}Medical} &
  Yearly &
  82.1074 &
  {\color[HTML]{343434} 2074.58} &
   1.8515   \\ \hline
\cellcolor[HTML]{FFFFFF} &
  Hourly &
  0.0024 &
  
  {\color[HTML]{343434} 0.0497} &
   20.6551   \\ \cline{2-5} 
\cellcolor[HTML]{FFFFFF} &
  Daily &
  0.0577 &
  {\color[HTML]{343434} 0.2704} &
   4.6881   \\ \cline{2-5} 
\cellcolor[HTML]{FFFFFF} &
  Weekly &
  0.4022 &
  {\color[HTML]{343434} 0.9922} &
  2.4672   \\ \cline{2-5} 
\cellcolor[HTML]{FFFFFF} &
  Monthly &
  1.7435 &
  {\color[HTML]{343434} 3.4492} &
   1.9783 \\ \cline{2-5} 
\multirow{-5}{*}{\cellcolor[HTML]{FFFFFF}Alarms} &
  Yearly &
  20.6866 &
  {\color[HTML]{343434} 37.0539} &
  1.7912   \\ \hline
\cellcolor[HTML]{FFFFFF} &
  Hourly &
  0.0006 &

  {\color[HTML]{343434} 0.0252} &
 40.1769    \\ \cline{2-5} 
\cellcolor[HTML]{FFFFFF} &
  Daily &
  0.015 &
  {\color[HTML]{343434} 0.1282} &
   8.5353  \\ \cline{2-5} 
\cellcolor[HTML]{FFFFFF} &
  Weekly &
  0.1048 &
  {\color[HTML]{343434} 0.3932} &
    3.7537  \\ \cline{2-5} 
\cellcolor[HTML]{FFFFFF} &
  Monthly &
  0.4542 &
  {\color[HTML]{343434} 1.1284} &
  2.4846    \\ \cline{2-5} 
\multirow{-5}{*}{\cellcolor[HTML]{FFFFFF}Citizen Assist} &
  Yearly &
  5.3885 &
  {\color[HTML]{343434} 10.2884} &
   1.9093  \\ \hline
\cellcolor[HTML]{FFFFFF} &
  Hourly &
  0.0013 &
  {\color[HTML]{343434} 0.0365} &
    28.0864  \\ \cline{2-5} 
\cellcolor[HTML]{FFFFFF} &
  Daily &
  0.0312 &
  {\color[HTML]{343434} 0.1841} &
   5.9065   \\ \cline{2-5} 
\cellcolor[HTML]{FFFFFF} &
  Weekly &
  0.2174 &
  {\color[HTML]{343434} 0.537} &
   2.4703  \\ \cline{2-5} 
\cellcolor[HTML]{FFFFFF} &
  Monthly &
  0.9423 &
  {\color[HTML]{343434} 1.4466} &
   1.5351  \\ \cline{2-5} 
\multirow{-5}{*}{\cellcolor[HTML]{FFFFFF}\begin{tabular}[c]{@{}c@{}}Motor \\ Vehicle \\ Incident\end{tabular}} &
  Yearly &
  11.181 &
  {\color[HTML]{343434} 12.7183} &
   1.1375 \\ \hline
\cellcolor[HTML]{FFFFFF} &
  Hourly &
  0.0001 &
  {\color[HTML]{343434} 0.0113} &
  89.1289  \\ \cline{2-5} 
\cellcolor[HTML]{FFFFFF} &
  Daily &
  0.0031 &
  {\color[HTML]{343434} 0.0567} &
  18.5966 \\ \cline{2-5} 
\cellcolor[HTML]{FFFFFF} &
  Weekly &
  0.0213 &
  {\color[HTML]{343434} 0.1643} &
   7.7254   \\ \cline{2-5} 
\cellcolor[HTML]{FFFFFF} &
  Monthly &
  0.0922 &
  {\color[HTML]{343434} 0.4335} &
   4.7010  \\ \cline{2-5} 
\multirow{-5}{*}{\cellcolor[HTML]{FFFFFF}Rescue} &
  Yearly &
  1.0941 &
  {\color[HTML]{343434} 3.6425} &
   3.3293  \\ \hline
\end{tabular}%
}
\end{table}

Figures ~\ref{fig-events-in-neighborhood} (a), (b), and (c) show heat-map plots of the rate of daily event occurrence of Fire, Medical, Alarms, Citizen Assist, Motor Vehicle (Traffic) incidents, and Rescue, respectively. (Table \ref{descriptive-statistics of events by neighborhood}) shows that a fire event incident happens every $2500$ hours (around 100 days), and a medical event every $105$ hours (4 days). However, in the central area neighborhoods, a fire event incident occurs every $8$ days, and $4$ medical event incidents occur every day. Therefore, we see that the occurrence rate varies widely by neighborhood. Fire events are more likely to happen in neighborhoods of the central city area. In contrast, the traffic incident events are scattered throughout the city. Most of the medical event incidents occurred in neighborhoods of the central core city. This suggests that each neighborhood may have different characteristics that impact the occurrence of different event types.

\begin{figure*} []

      \centering
        \includegraphics[width=\textwidth,keepaspectratio]{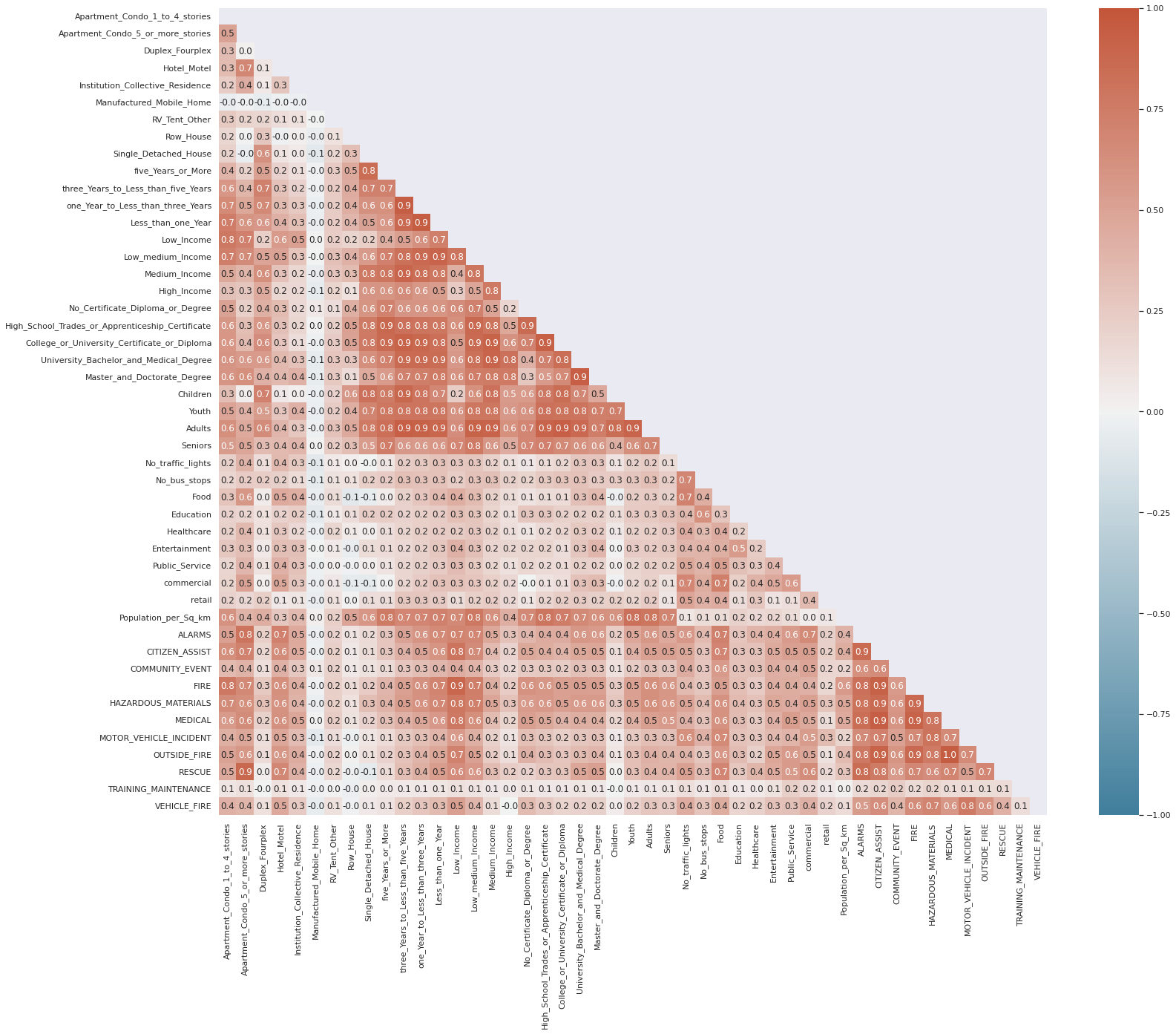}
    \caption{\small {\bf Correlation coefficients between feature and event}}
   \label{correlation-heatmap}
    \end{figure*}

\begin{figure*} []
		\centering

		\subfloat[Fire]{
			\label{imp_var_fr}
			\includegraphics[width=0.48\textwidth, height=0.40\textwidth]{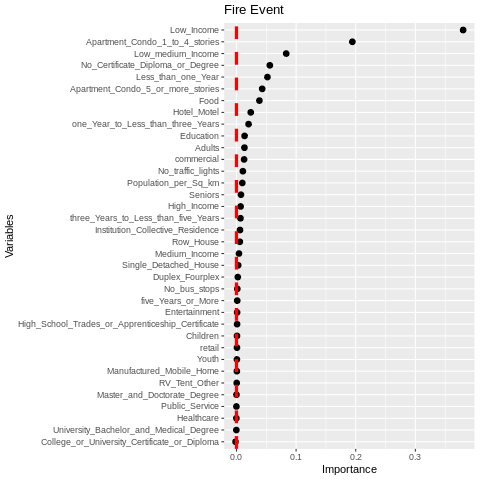}}
        \subfloat[Medical]{
			\label{imp_var_md}
			\includegraphics[width=0.48\textwidth, height=0.40\textwidth]{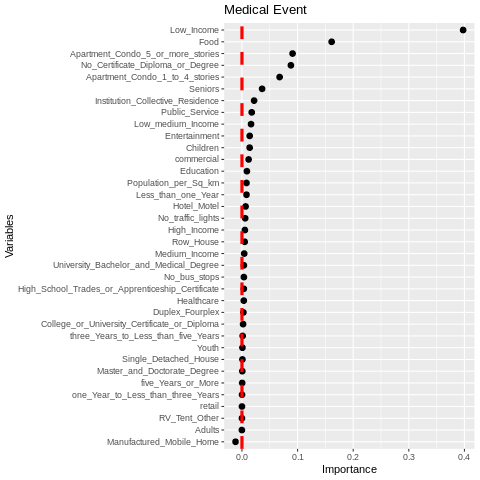}}
			
		\subfloat[Alarms]{
			\label{imp_var_al}
			\includegraphics[width=0.48\textwidth, height=0.40\textwidth]{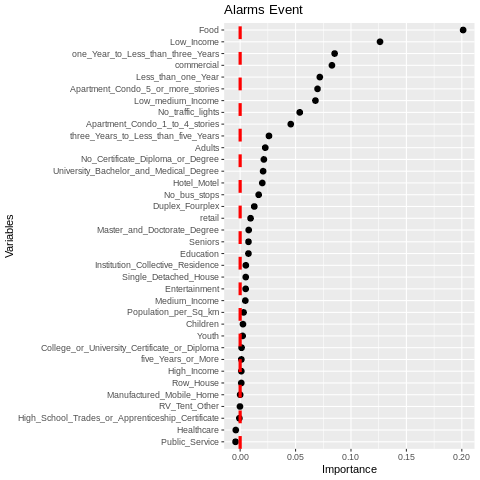}}
		\subfloat[Citizen Assist]{
			\label{imp_var_ca}
			\includegraphics[width=0.48\textwidth, height=0.40\textwidth]{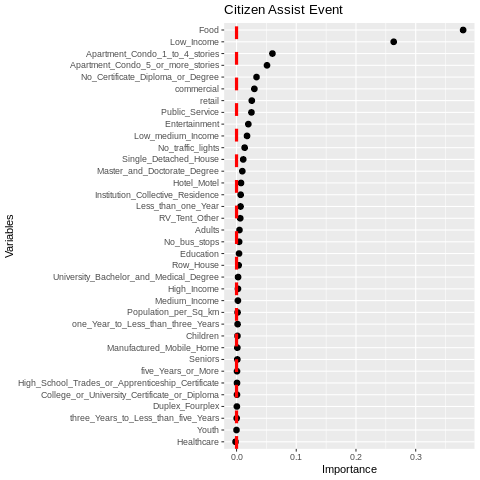}}
			
		\subfloat[Motor Vehicle Incident]{
			\label{imp_var_ta}
			\includegraphics[width=0.48\textwidth, height=0.40\textwidth]{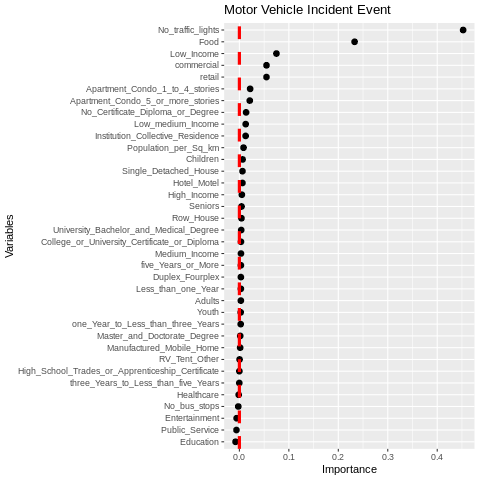}}
		\subfloat[Rescue]{
			\label{imp_var_rc}
			\includegraphics[width=0.48\textwidth, height=0.40\textwidth]{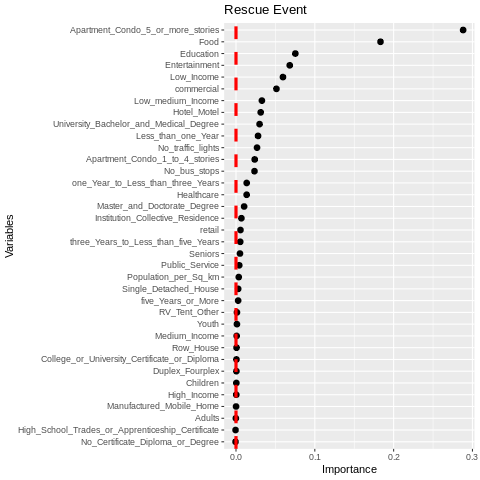}}

\caption {\small {Results of Feature Importance Analysis with Conditional Random Forest Model for (a) Fire, (b) Alarms, (c) Medical, (d) Motor Vehicle Incident, (e) Citizen Assist, and (f) Rescue events}}
\label{fig_imp_variable}

\end{figure*}

\section{Feature Analysis \& Selection} \label{sec-feature-analysis}

We now present a correlation analysis of features and event types to help us determine how a predictor variable (feature-independent variable) is associated with the event (dependent variable). We expect that a model with strongly associated predictor variables will make good predictions. We employed two methods of analysis:  Pearson Correlation Coefficient, and Conditional Random Forest. We analyzed the correlation coefficient of all the features with the events (dependent variable) using the Pearson correlation coefficient. To identify the feature set for predicting the events, we used the Conditional Random Forest (CRF) model. 

\subsection{Correlation Analysis}
\label{label:correlation-analysis}
A correlation coefficient measures the degree of association between two variables. We analyzed the correlation coefficient between the predictor variable (i.e., feature) and the response/target variable (i.e., event). 
Let $x_i$ and $y_i$ denote static features and events, respectively. The Pearson correlation coefficient of features (x) and event (y) denoted by $\rho$ is defined as follows:

\begin{equation}
\rho = \frac{\sum_{i}^{n} (x_i- \bar{x}) (y_i- \bar{y})}{\sqrt{\sum_{i}^{n}(x_i-\bar{x})^2}\sqrt{\sum_{i}^{n}(y_i-\bar{y})^2}} 
\end{equation}

A positive correlation indicates that features and events tend to move in the same direction. A negative correlation indicates features and events tend to move in opposite directions. Figure~\ref{correlation-heatmap} shows the correlation coefficients of features and events (rounded to a single decimal place). We see a strong association between household income, population, and residential building types with each of the event types.  Food, a point of interest-related feature, has a significant association with the occurrence of medical, alarms, citizen-assist, motor vehicle incidents, and rescue events. The number of hotels or motels is associated with medical, alarms, citizen assist, and rescue events. The number of traffic lights is only associated with the motor vehicle incident event. Similarly, the length of term residency feature of less than one year has a good correlation with fire events. 

\subsection{Feature Importance Analysis}
\label{label:feature-importance}

A conditional Random Forest (CRF)~\cite{Hothorn2006} analysis was conducted to identify the important features (i.e., predictors) associated with EFRS event(s) (i.e., response/target variable) for the model development. The CRF method involves developing conditional interface trees (e.g., event trees) by repeatedly and randomly subsetting data and identifying the features that are most associated with events in the subsets. The conditional distribution of the statistics is derived to measure the association between the target variable (i.e., event) and the predictor variables (e.g., features). A statistical test with $p$-values is used to determine whether there exist statistically significant associations between any of the features (predictors) and event (i.e., target/response) variables. The most strongly associated feature with the event (target variable) is selected.

We conducted several experiments to identify the important features (i.e., predictor variables) for the predictive model development. The package “partykit\cite{Hothorn2015}” in R was used for this analysis. This analysis
developed 1000 trees (event trees) for each event type; the collective output from
those 1000 trees produced the final importance ranking of the
features (i.e., predictor variables) for each event type. In this study, we have considered the six common EFRS events: Fire, Alarms, Medical, Motor Vehicle Incident, Citizen Assist, and Rescue events. Figure~\ref{fig_imp_variable} shows the important features of the six event categories. The features with the high score from the top are the important features. Each event type has a different important feature list because of their different characteristics. Low-income, food and apartments or condominiums with 5 or more stories are identified as important features of most of the event types. A list of importance features for each of the event types with score threshold $\ge 0.05$ is presented in Table ~\ref{importance-variables}. The findings based on both correlation and CRF analysis presented in this section are summarized as follows:
\begin{itemize}
    \item There is a strong association between events and low household income and residential building types. The event incidents increase as the number of the low-income population, condos/apt, and hotels/motels increase.
    \item The point of interest (PoI) related feature such as the number of restaurants and bars/cafes/food courts (food) has a significant association with the occurrence of events.
    \item Shorter length of residency term is associated with fire events. 
    \item The commercial building types (commercial or retail) are also associated with the occurrence of most of the event types.
    \item Road traffic information including the number of traffic lights is significantly associated with motor vehicle incident events.
\end{itemize}

\begin{table*} []
\centering
\caption{ List of importance features with score $\geq 0.05$}.
\label{importance-variables}
\begin{tabular}{|l|l|}
\hline
\rowcolor[HTML]{EFEFEF} 
\textbf{Event Types} &
  \textbf{Importance Features  (Variables)} \\ \hline
Fire (FR) &
  \begin{tabular}[c]{@{}l@{}}Low\_Income, Apartment\_Condo\_1\_to\_4\_stories, Low\_medium\_Income ,  No\_Certificate\_Diploma\_or\_Degree , \\ Less\_than\_one\_Year \end{tabular} \\ \hline
Medical (MD) &
  \begin{tabular}[c]{@{}l@{}}Low\_Income,  Food,  Apartment\_Condo\_5\_or\_more\_stories , No\_Certificate\_Diploma\_or\_Degree,\\ Apartment\_Condo\_1\_to\_4\_stories \end{tabular}\\ \hline
Alarms (AL) &
  \begin{tabular}[c]{@{}l@{}}Food, Low\_Income, one\_Year\_to\_Less\_than\_three\_Years, commercial, Less\_than\_one\_Year,  \\ Apartment\_Condo\_5\_or\_more\_stories, Low\_medium\_Income,  No\_of\_traffic\_lights \end{tabular} \\ \hline
\begin{tabular}[c]{@{}l@{}}Citizen \\ Assist (CA)\end{tabular} &
  \begin{tabular}[c]{@{}l@{}}Food,  Low\_Income, Apartment\_Condo\_1\_to\_4\_stories, Apartment\_Condo\_5\_or\_more\_stories \end{tabular} \\ \hline
\begin{tabular}[c]{@{}l@{}}Motor Vehicle \\ Incident (TA) \end{tabular} &
  \begin{tabular}[c]{@{}l@{}}No\_of\_traffic\_lights, Food, Low\_Income, commercial, retail \end{tabular} \\ \hline
Rescue (RC) &
  \begin{tabular}[c]{@{}l@{}}Apartment\_Condo\_5\_or\_more\_stories, Food,  Education, Entertainment, Low\_Income, commercial \end{tabular} \\ \hline
\end{tabular}%
\end{table*}

\section{Model Development} \label{exerimental-setup}
We now present the predictive model development for EFRS event prediction and the associated experimental setup.

\subsection{Regression Model}  \label{sec-nb2-regression}
We developed a regression model to predict EFRS events using the Negative Binomial (NB2) model. Negative binomial regression is commonly used to test for the connections between confounding (events) and predictor variables (features) on over-dispersed count data. This fits EFRS events because the event counts are over-dispersed. The NB2 model is based on a Poisson-gamma mixture distribution where the Poisson heterogeneity is modeled using a gamma distribution~\cite{Hilbe2011,Cameron2013}. The NB2 distribution is a two-parameter model with mean $\mu$ and dispersion $\alpha$. The variance of the NB2 distribution is estimated using the parameters $\mu$ and $\alpha$. 

Let $y$ denotes a vector of EFRS event counts seen for a given time duration for a given event type. $X$ denotes a matrix of predictor/regressor variables (e.g., static features), and $y_i$ is the number of events in the ith observation. The mean of $y$, $\mu$ is determined in terms of exposure time $t$ and a set of $k$ regressor variables (the $x$’s). The mean and variance of each $y_i$ can be defined as follows: 
\begin{equation}
\label{eq-regression}
          \mu_i= exp(\ln (t_i) +\beta_1 x_{1i}+ \beta_2 x_{2i} +\dots \beta_k x_{ki})
\end{equation} 
\begin{equation}
          var(y_i)= \mu_i+ \alpha \mu_i^2 
\end{equation}
    where, $t_i$ represents the exposure time for a particular event (e.g. hour, day, week, month, year), and $\beta_1$, $\beta_2$, $\beta_3$, \dots, $\beta_k$ are the (unknown) regression coefficients which are to be estimated with the training data. The maximum likelihood method is used to estimate the regression coefficients~\cite{Cameron2013}. Their estimates are denoted by $b_1$, $b_2$, \dots, $b_k$. The parameter $\mu$ is a vector of the event rates which can be estimated using the Poisson model. Each $\mu_i$ corresponds to the event count $y_i$ in the vector $y$. Each $\mu_i$ can be estimated using the Poisson model. The parameter $\alpha$ is estimated using ordinary least square regression with $y_i$, $i^{th}$ outcome, and $\mu_i$ is $i^{th}$ rate. The probability distribution function of the NB2 model for the  $i^{th}$ observation ( i.e., $y_i$ event ) is defined using Eq.~\ref{eq-prob}, where the function $\Gamma (\cdot) $ is the gamma function~\cite{Cameron2013}.
\begin{equation}
\label{eq-prob}
\small {
    Pr\left({y}_{i}|{\mu}_{i}, \alpha \right)=\frac{\Gamma \left({y}_{i}+{\alpha }^{-1}\right)}{\Gamma \left({y}_{i}+1\right)\Gamma \left({\alpha }^{-1}\right)}{\left(\frac{1}{1+\alpha {\mu}_{i}}\right)}^{{\alpha }^{-1}}{\left(\frac{\alpha {\mu}_{i}}{1+\alpha {\mu}_{i}}\right)}^{{y}_{i}}
    }
\end{equation}


The performance of the regression models can be evaluated with regression metrics such as Mean Absolute Error (MAE), Root Mean Squared Error (RMSE). These metrics measure the difference between the actual values and the predicted values on the regression line. MAE and RMSE evaluation metrics are given by Eq. \ref{mae}, and Eq. \ref{rmse}. These equations illustrate the calculation of MAE and  RMSE metrics, where $y_i$ is the $i^{th}$ actual value, and $\hat{y_i}$ is the corresponding predicted value. The percentage error (\% Error) for each model is another metric to assess the performance of the models.
\begin{equation}
   MAE =\sum_{i=1}^{n} \frac{|y_i- \hat{y}_i|}{n}
 \quad \label{mae}  
\end{equation}

\begin{equation}
   RMSE=\sqrt{\sum_{i=1}^{n} \frac{(y_i- \hat{y}_i)^2}{n}} 
  \quad \label{rmse}  
\end{equation}

\subsection{Experimental Setup}
\label{sec-experimental-setup}

We conducted experiments with regression models for neighborhood and fire-station regions. The prediction of individual event types by neighborhood proved very difficult because of their low event occurrence rate. Therefore, we focused on weekly and monthly predictive models for total event incident counts using the union of all the important predictor variables of event types. Eq.~\ref{eq-events} shows a regression model for the total number of event counts (Events) with the feature sets (predictors) of neighborhoods, where each $b_i$, $i=0,1,2,\dots$, represents the coefficient of regression variable.

\begin{equation}
\resizebox{0.90\hsize}{!}{$%
    Y\left (Events\right) = \exp \left[ \begin{array}{r}
        b_0 + b_1 \times Food + b_2 \times Low\_Income  \\
        + b_3 \times retail + b_4 \times commercial     \\ 
        + b_5\times Education + b_6\times Entertainment \\
        + b_7 \times one\_Year\_to\_Less\_than\_three\_Years \\ 
        + b_8 \times  Apartment\_Condo\_5\_or\_more\_stories \\ 
        + b_9 \times Apartment\_Condo\_1\_to\_4\_stories \\
        + b_{10}\times No\_Certificate\_Diploma\_or\_Degree \\
        + b_{11} \times Low\_medium\_Income \\
        + b_{12}\times No\_traffic\_lights \\ 
        + b_{13} \times Less\_than\_one\_Year 
\end{array} \right]
$%
}
\label{eq-events}
\end{equation}

For the fire station region prediction, we conducted experiments with different regression models for each event type. Based on the model developed in Section~\ref{sec-nb2-regression}, regression models were formulated for each event type considering the important predictor variables identified from the feature importance analysis in Section~\ref{label:feature-importance}, and shown in Table~\ref{importance-variables}. We formulated models for the event types: fire, medical, alarms, citizen-assist, motor-vehicle incident, and rescue events. For example, Eq.~\ref{eq-fr} shows a model expression for the Fire event. Each $b_i$, $i=0, 1, 2, \dots$, represents the coefficient of regression variables, which were estimated using model training. 

\begin{equation}
\resizebox{0.90\hsize}{!}{$%
    Y\left (FR\right) = \exp \left[ \begin{array}{r}
    b_0 + b_1 \times Low\_Income \\ 
    + b_2 \times Low\_medium\_Income \\ 
    + b_3 \times  Less\_than\_one\_Year \\
    + b_4 \times Apartment\_Condo\_1\_to\_4\_stories \\ 
    + b_5 \times No\_Certificate\_Diploma\_or\_Degree
\end{array} \right]
$%
}
\label{eq-fr}
\end{equation}

Statsmodels~\cite{seabold2010statsmodels, stat_model_api} APIs were used to build the models and perform the experiments. We performed experiments for training, prediction, and evaluation of weekly and monthly durations at different spatial ( fire-station regions and neighborhoods) resolutions of the city.

\section{Event Prediction by Neighborhood}
\label{prediction-by-neighborhoods}

In this section, we present the event prediction results by neighborhood including performance analysis of the models and risk classification of the neighborhoods. We conducted several experiments to train models and made the event predictions with neighborhood features for each neighborhood. We build a generic predictive model with the identified important variables for forecasting the occurrence of the events in each neighborhood. 

\begin{figure*} [htb!]
		\centering
		\subfloat[]{%
			\includegraphics[width=0.5\textwidth, height=\textwidth,keepaspectratio]{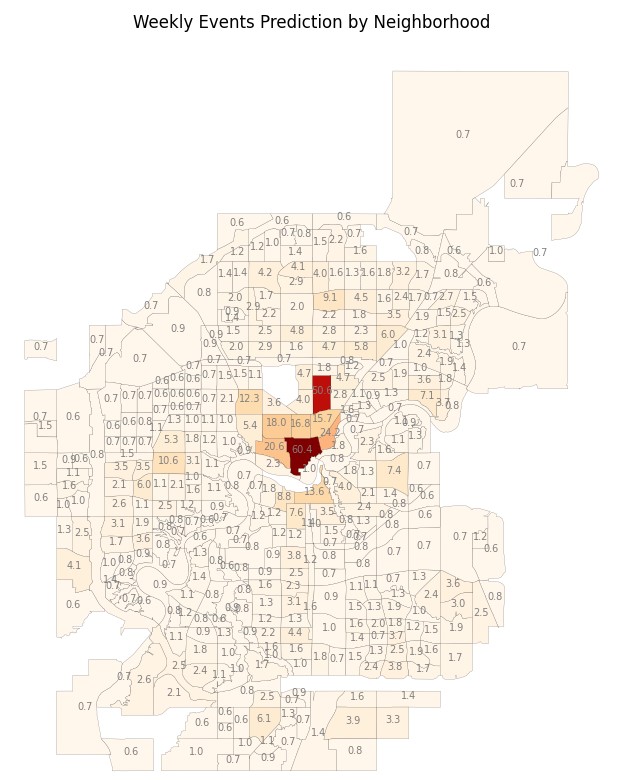}}
        \subfloat[]{%
			\includegraphics[width=0.5\textwidth, height=\textwidth,keepaspectratio]{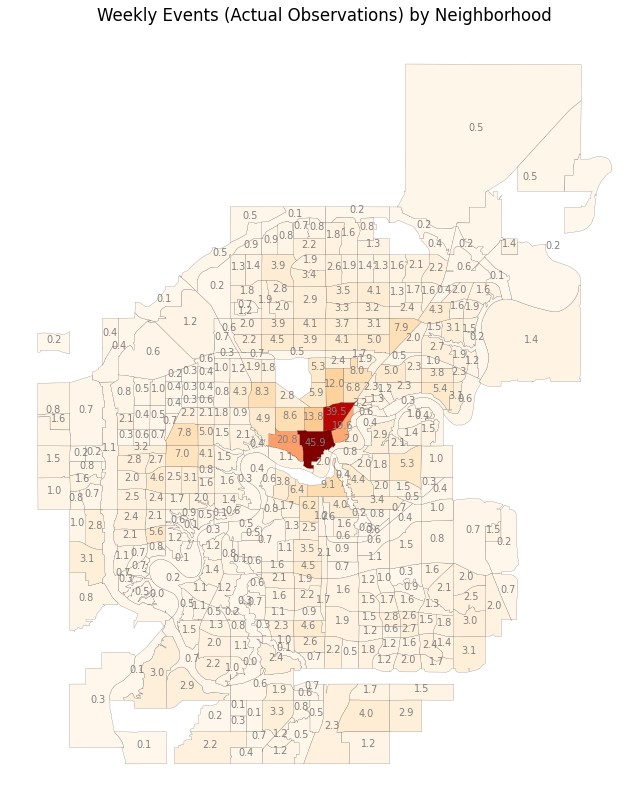}}		
\caption {\small {Weekly Event Predictions by Neighborhood: (a) Predicted Events (b) Actual Events. The annotation number on the neighborhood polygon map represents a mean of weekly predicted/actual events. Neighborhoods with dark (red) color have more event incidents.}}
\label{weekly_by_neighborhood}
\end{figure*}

\subsection{Weekly Event Predictions} 

In this experiment, we trained and evaluated the weekly predictive model with the neighborhood's demographic, socioeconomic, and map-related (PoI) features and weekly event data from 2011-01-03 to 2022-08-01. We randomly split the data into training and test samples with 70\% of data for training and 30\% of data for testing, resulting in  161189 and 68781 samples, respectively. We forecast the weekly event occurrences in each neighborhood using the neighborhood's features. The predictions were validated with out-sample test data. Figure \ref{weekly_by_neighborhood} compares the weekly event count predicted results for each neighborhood with the actual event observations. Weekly event predictions with static features show a good performance for forecasting the event occurrence except for a few special neighborhoods in the downtown core. These Weekly predictions for each neighborhood can be used in weekly emergency preparedness and planning.

\begin{figure*} [htb!]
		\centering
		\subfloat[]{
			\label{}
			\includegraphics[width=0.5\textwidth, height=\textwidth,keepaspectratio]{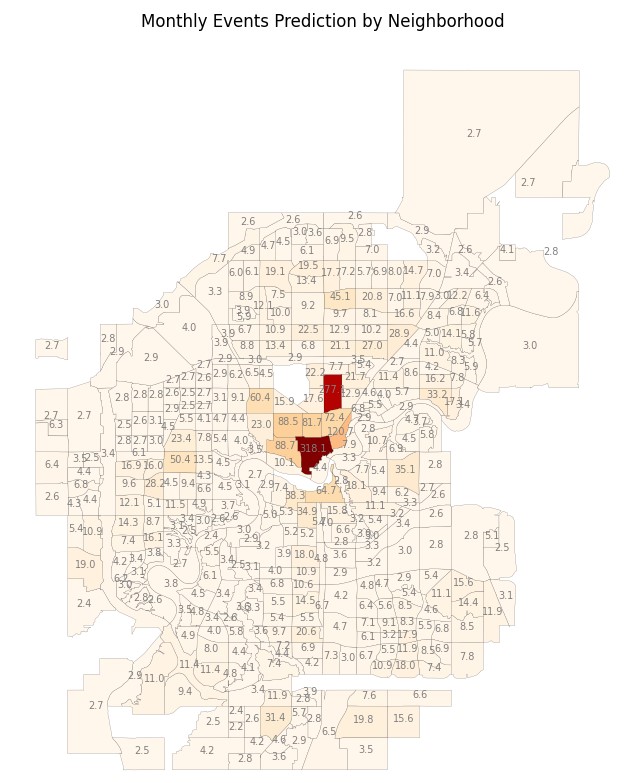}}
        \subfloat[]{
			\label{}
			\includegraphics[width=0.5\textwidth, height=\textwidth,keepaspectratio]{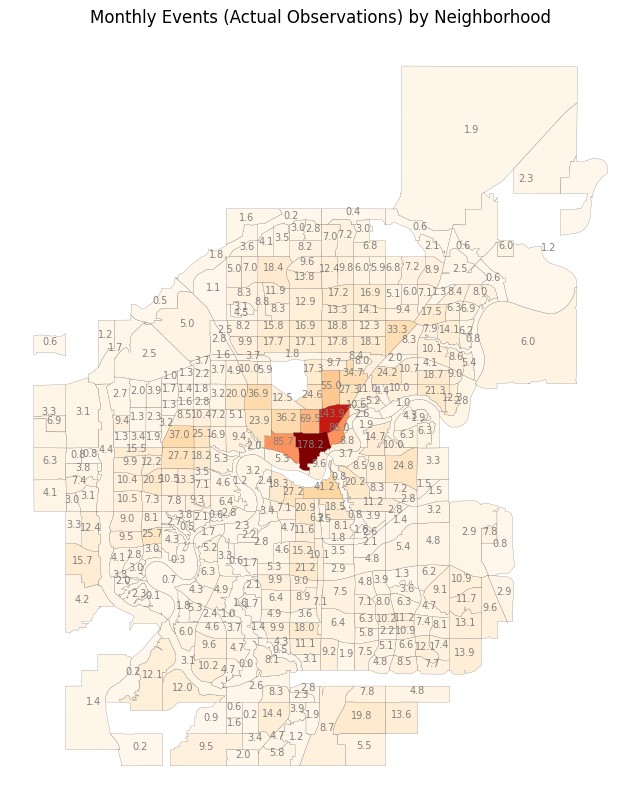}}
			
\caption {\small {Monthly Event Predictions by Neighborhood: (a) Predicted Events (b) Actual Events. The annotation number on the neighborhood polygon map represents a mean of monthly predicted/actual events. Neighborhoods with dark (red) color have more event incidents.}}
\label{monthly_by_neighborhood}
\end{figure*}

\subsection{Monthly Event Predictions}

We also conducted experiments to predict monthly event counts with the static features of the neighborhood. As in the previous section, we trained the models with monthly event data and then validated the predictions with out-sample test data. The monthly event data with static features contains 36729 training and 16051 testing samples. Figure \ref{monthly_by_neighborhood} compares the mean of the monthly predictions with the mean of the actual event incidents in each neighborhood. The monthly predictions by neighborhood are relatively good as compared to the weekly predictions. However, the prediction for some neighborhoods have a relatively high error possibly because these neighborhoods have special characteristics. 
These monthly predictions can be used by the city for making prevention and mitigation plans at a neighborhood level.

\subsection{Model Evaluation \& Prediction Error Analysis}

We evaluated the model's performance with regression metrics and residual distribution analysis. The overall performance of the weekly and monthly predictive models was evaluated with mean absolute error (MAE) and root mean square error (RMSE) regression metrics. With weekly predictions, MAE and RMSE are 0.81 (mean of actual weekly observation is 2.06) and 2.59, respectively. Similarly, MAE and RMSE with monthly predictions are 4.09 (mean of actual monthly observation is 8.88), and 15.16, respectively. 

\begin{figure*} [hbt!]
		\centering
		\subfloat[]{
			\label{}
			\includegraphics[width=0.5\textwidth,keepaspectratio]{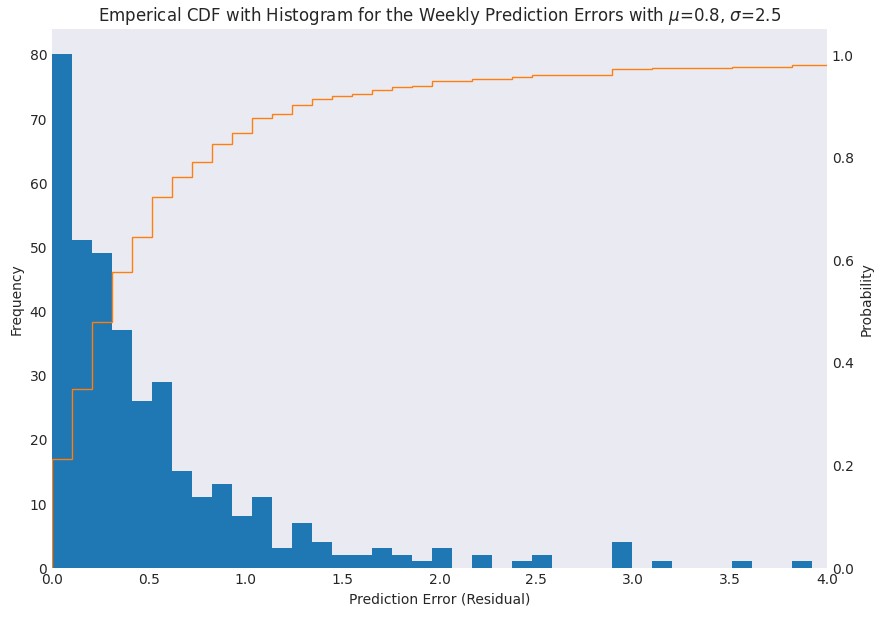}}
        \subfloat[]{
			\label{}
			\includegraphics[width=0.5\textwidth,keepaspectratio]{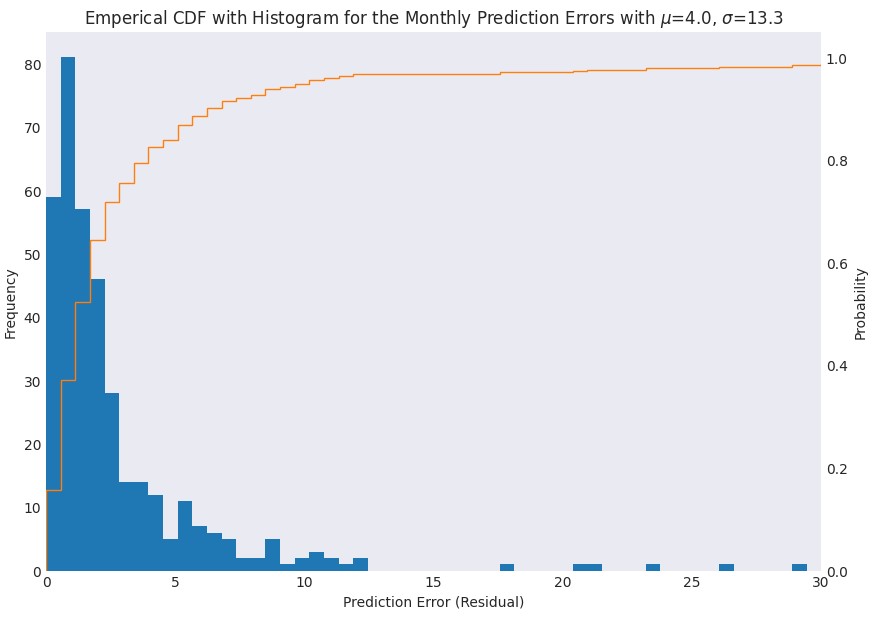}}
			
\caption {\small {Empirical CDF with Histograms of Prediction Errors: (a) Weekly Prediction Errors (b) Monthly Prediction Errors.}}
\label{Error_by_neighborhood}
\end{figure*}

We also examined the distribution of the prediction error (residual) for each neighborhood using an empirical cumulative distribution function (CDF) of the test sample data. Figure \ref{Error_by_neighborhood} shows the empirical CDF plot with a histogram of the distribution of the errors of the predictive models. Figure \ref{Error_by_neighborhood} (a) shows a distribution of the weekly prediction errors with histogram. The results show that the model performs well except for a few neighborhoods. With this weekly prediction model, 47\% of the neighborhoods (175 out of 377) have 0 residual error, 42\% of the neighborhoods (160 out of 377) have error 1 (1 more or less event predicted), 5\% of the neighborhoods have error 2, and $\approx 90\%$ of the neighborhoods are with $\leq 1$ error. Similarly, \ref{Error_by_neighborhood} (b) shows a distribution of monthly prediction errors with histogram. Results show that this model also performs well except for a few neighborhoods. With this monthly prediction model, 16\% of the neighborhoods (60 out of 377) have 0 residual error, 30\% of the neighborhoods (110 out of 377) have 1 error, and overall, 95\% of the neighborhoods are with $\leq 10$ error.

\subsection{Risk Classification of Neighborhoods}

\begin{figure*} [htb!]
		\centering
		\subfloat[]{
			\includegraphics[width=0.5\textwidth, height=\textwidth,keepaspectratio]{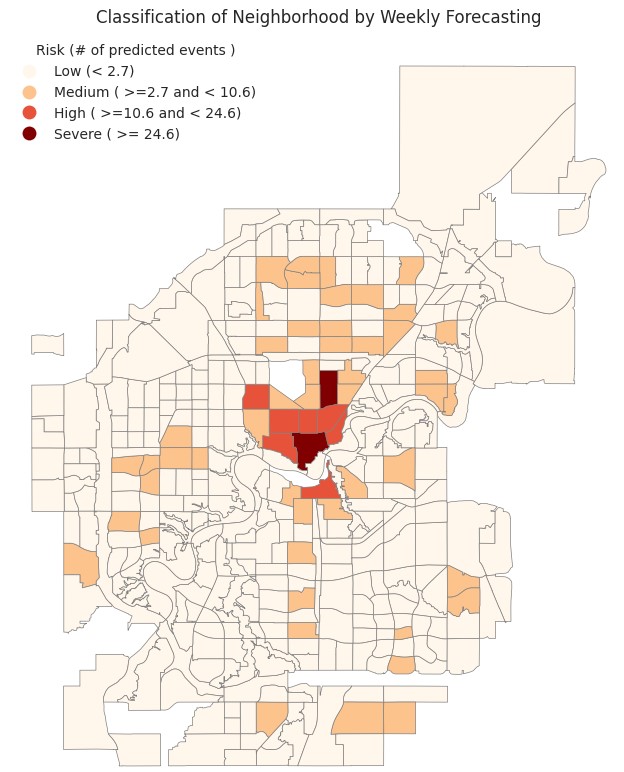}}
        \subfloat[]{
			
			\includegraphics[width=0.5\textwidth, height=\textwidth,keepaspectratio]{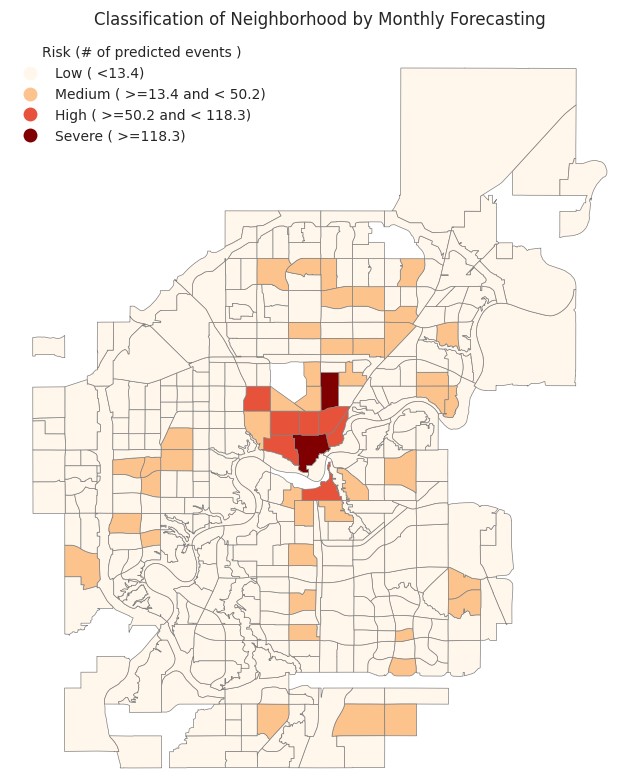}}
			
\caption {\small {Classifying the Neighborhoods by (a) Weekly Prediction; and (b) Monthly Prediction.}}
\label{classification_of_neighborhood}
\end{figure*}

\begin{figure*} [!hbt]
		\centering
		\subfloat[]{
			\label{fr_by_frs2}
			\includegraphics[width=0.5\textwidth, height=0.48\textwidth,keepaspectratio]{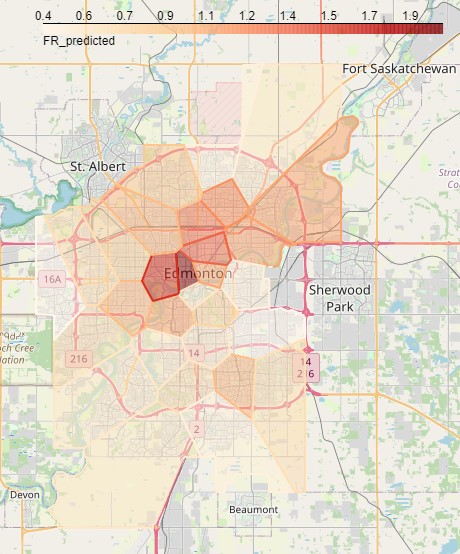}}
        \subfloat[]{
			\label{all_by_frs2}
			\includegraphics[width=0.5\textwidth, height=0.48\textwidth,keepaspectratio]{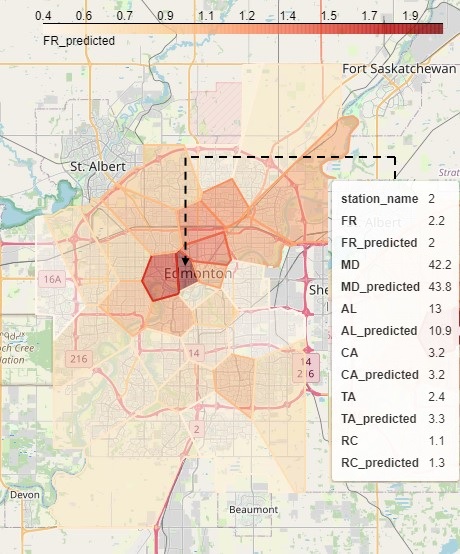}}
			
\caption {\small {Weekly Event Predictions by Fire Stations: (a) Fire Events Predictions, (b) Predicted and Actual Events for Fire Station 2.}}
\label{fig_fire_prediction_by_fr_stations_weekly}
\end{figure*}

We classified the neighborhoods based on the frequency of the predicted events in weekly and monthly time intervals. We define risk as the likelihood of occurrence of an event in a certain time period. We quantitatively expressed it into four risk classes low, medium, high, and severe based on the frequencies and applied Jenks' Natural Breaks \cite{Jenks1967data} method for the classification. Jenks' Natural Breaks is a data classification method suitable for data that have a high variance. This method seems useful to classify and visualize the event predictions made by our predictive models as we have predictions with high variations.  

We classified neighborhoods based on the forecasting values using the Jenks' Natural Breaks. We define four risk classes with low, medium, high. Figures~\ref{classification_of_neighborhood} (a) and ~\ref{classification_of_neighborhood} (b) present a risk class for each neighborhood based on the weekly and monthly predicted values, respectively. Both results show that the neighborhoods with numbers 1010 and 1090 have severe risk as each of these neighborhoods has an occurrence rate of 24.6 or more per week and 118.3 or more events per month. Neighborhoods 1020, 1030, 1140, 1150, 1180, 3240, 4180, and 5480 have a high risk (weekly 10.6 to 24.6 events predicted), which is reasonable since these neighborhoods belong to the central city area.  This risk classification can help the city make neighborhood preparedness plans to help prevent and reduce the impact of event incidents.

\section{Event Prediction by Fire-station}
\label{sec-predictions-by-fire-station}
We conducted several experiments to train models to predict event counts by Fire-station using the same methodology as for neighborhoods.

\begin{figure*} [!hbt]
\centering
 \includegraphics[width=0.95\textwidth,keepaspectratio]{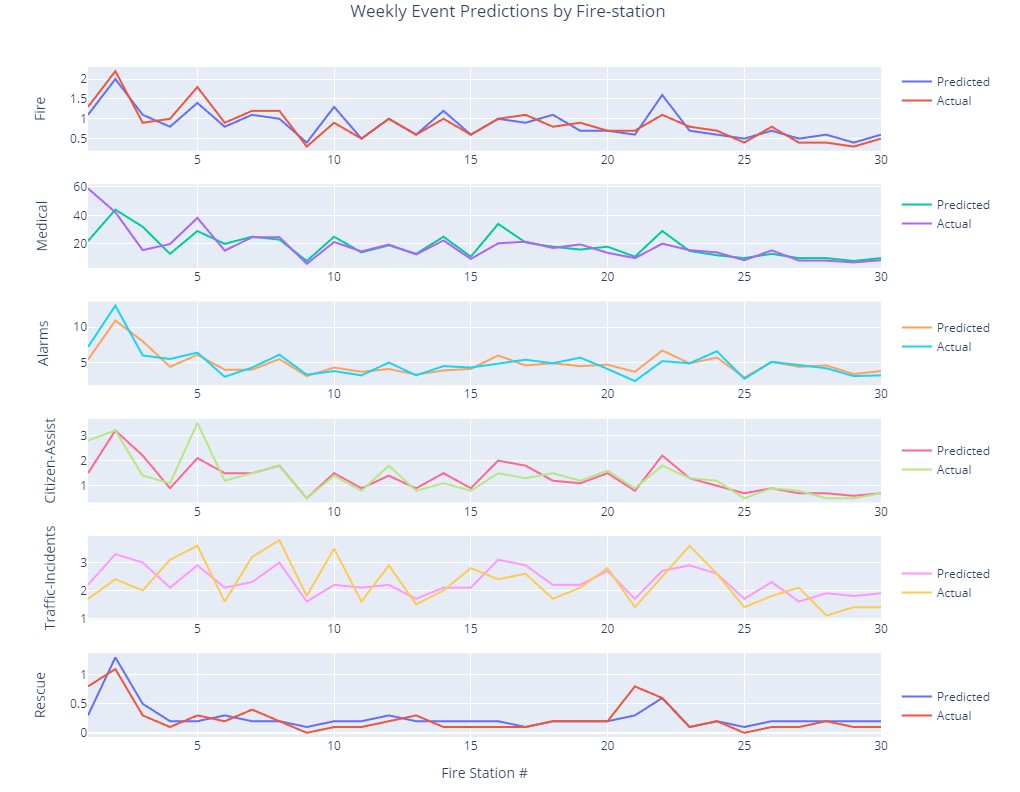}
\caption {\small {Comparative Analysis of Weekly Event Predictions with Actual Events.
}}
\label{fig_fire_prediction_by_fr_station_weekly2all}
\end{figure*}

\subsection{ Weekly Event Predictions}

We conducted experiments to predict weekly events with the static features of Fire stations. In these experiments, we trained and evaluated the models with demographic, map-related (PoI) features, and weekly event data from 2011-01-03 to 2022-08-01. We randomly split the data into training and test samples with 70\% of data for training and 30\% of data for testing. We made the predictions for each fire station and validated them with out-sample test event data. The training and test data consisted of 12721 and 5422 samples, respectively.

Figure~\ref{fig_fire_prediction_by_fr_stations_weekly} shows heap-map plots of weekly prediction and actual results. Figure~\ref{fig_fire_prediction_by_fr_stations_weekly} (a) shows that there is a higher chance for fire events to occur mainly in two fire stations in the central part of the city. Figure~\ref{fig_fire_prediction_by_fr_stations_weekly} (b) shows the actual and predicted number of events for different event types in the fire station 2. The polygon boundaries of each fire station are derived from the spatially nearest station. Actual zones would be impacted by accessibility and the further refinement of the station-level polygons could strengthen proper attribution of emergency event workload predictions.

Figure~\ref{fig_fire_prediction_by_fr_station_weekly2all} compares the mean of weekly event predictions for each fire station with the mean of the weekly actual event counts. The results show that weekly event predictions with the static features for each fire station exhibit a good performance for predicting the event occurrences except for a few special fire stations, stations 1 for medical and 5 for fire events which have a relatively high error. These two stations seem to have specific features that the model does not capture fully. There are many encampment and overdose incident calls to fire station 1 and such calls increase the frequency of medical events. Similarly, station 5 also has an interesting characteristic, and specifically for fire events. There are many points of interest (PoIs) such as stadiums, transit, and recreation centers within the areas of station 5, which would possibly increase the fire event incidents. This model did not fit well the rescue events because these events occur rarely and randomly. The weekly predictions made for each fire station with the given features of a fire station's region would help for weekly emergency preparedness and planning.

\begin{figure*} [hbt]
		\centering

		\subfloat[]{
			\label{fr_by_frs2}
			\includegraphics[width=0.5\textwidth, height=0.48\textwidth,keepaspectratio]{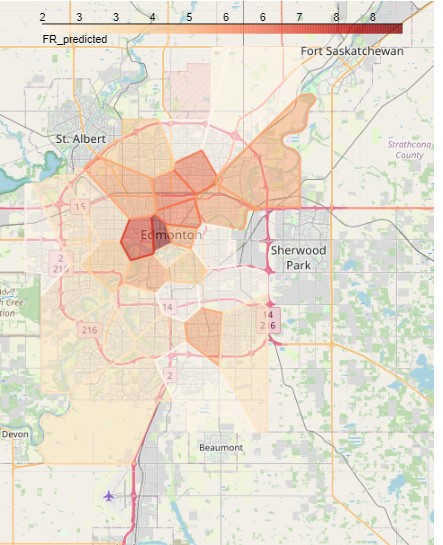}}
        \subfloat[]{
			\label{all_by_frs2}
			\includegraphics[width=0.5\textwidth, height=0.48\textwidth,keepaspectratio]{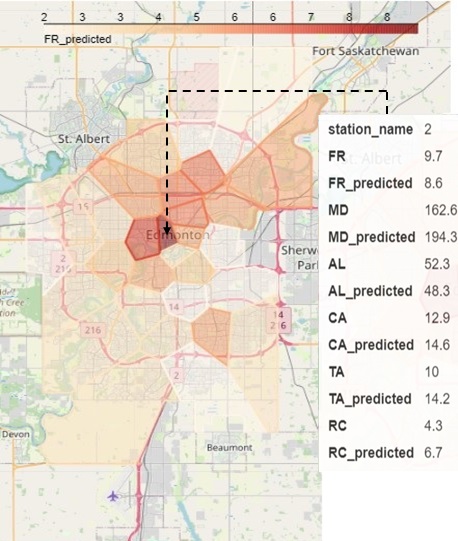}}
			
\caption {\small {Monthly Event Predictions by Fire Stations: (a) Fire Events Predictions, (b) Predicted and actual events for a Fire Station 2.}}
\label{fig_fire_prediction_by_fr_station_monthly}
\end{figure*}

\subsection{Monthly Event Predictions}

We conducted experiments to predict monthly events with the static features of the fire stations using the same methodology as in the previous section. We made the predictions for each fire station and then validated the predictions with out-sample test monthly event data. The monthly event data with static features contains 2933 training and 1237 testing samples. 

\begin{figure*} [hbt]
		\centering
 \includegraphics[width=1.0\textwidth,keepaspectratio]{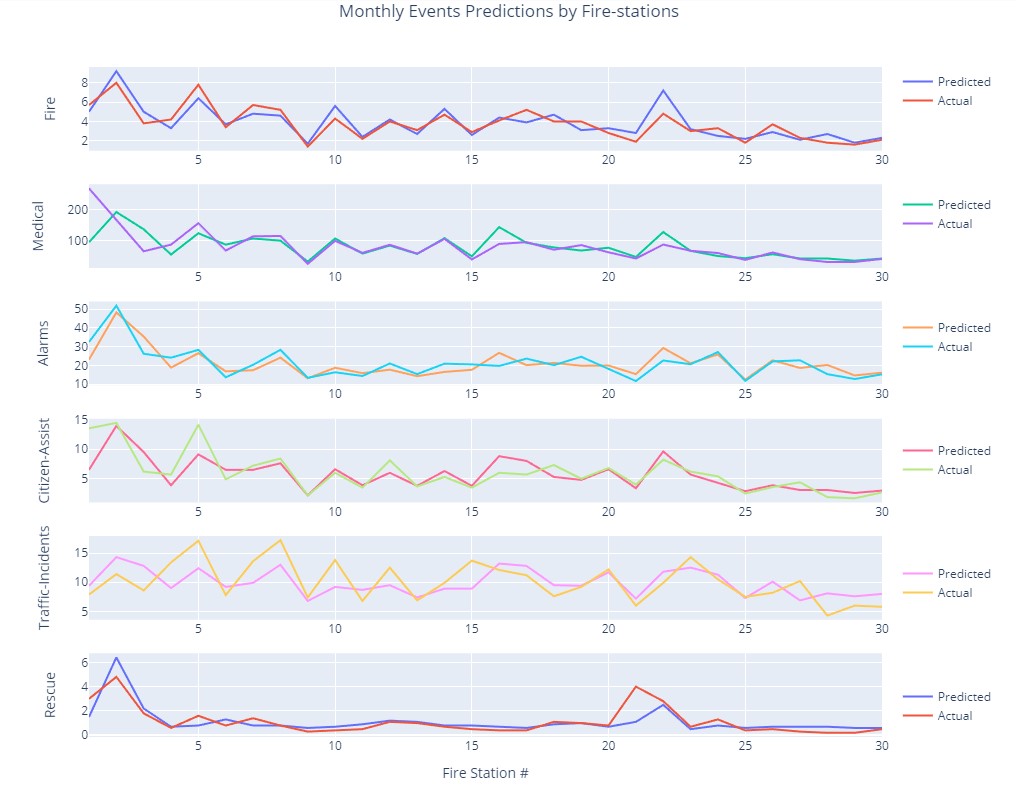}
\caption {\small {Comparative Analysis of Monthly Event Predictions with Actual Events. The X-axis and Y-axis represent fire station number and count of six different event types, respectively.}}
\label{fig_fire_prediction_by_fr_station_monthly2all}
\end{figure*}

Figure~\ref{fig_fire_prediction_by_fr_station_monthly} shows the heap-map plot of prediction results of the monthly event prediction models. Similarly, Figure~\ref{fig_fire_prediction_by_fr_station_monthly2all} compares the mean of the monthly predictions with the mean of the actual event incidents for each fire station. The monthly predictions by the fire station are relatively good in comparison to the weekly predictions. However, the prediction for stations 1, 5, and 21 have a high error because we developed a single model for all stations, but these stations have special characteristics. Fire station 21 has many rescue events because this station is along the river. This station is unique in that it is the only station without a pumper unit. Instead, it houses just a Rescue, with staff trained for swift water rescue boat response. The station location is not associated with event density, but instead river access. This impacts the call volume as well as call characteristics. As noted, they have a higher concentration of rescue events due to their proximity to the river, associated bridges, and surrounding parkland. Similarly, station 5 has relatively more fire events as there are many PoIs such as stadium, transit hub, recreation, \textit{etc.}  We can improve the predictions of the model for these stations by exploring other explanatory factors impacting the occurrence of these events in these stations.  The predictions made for each fire station with the given demographic and map-related features of the fire station's service area help with monthly emergency preparedness and planning.

\begin{figure*} [hbt!]
		\centering
 \includegraphics[width=1.0\textwidth,keepaspectratio]{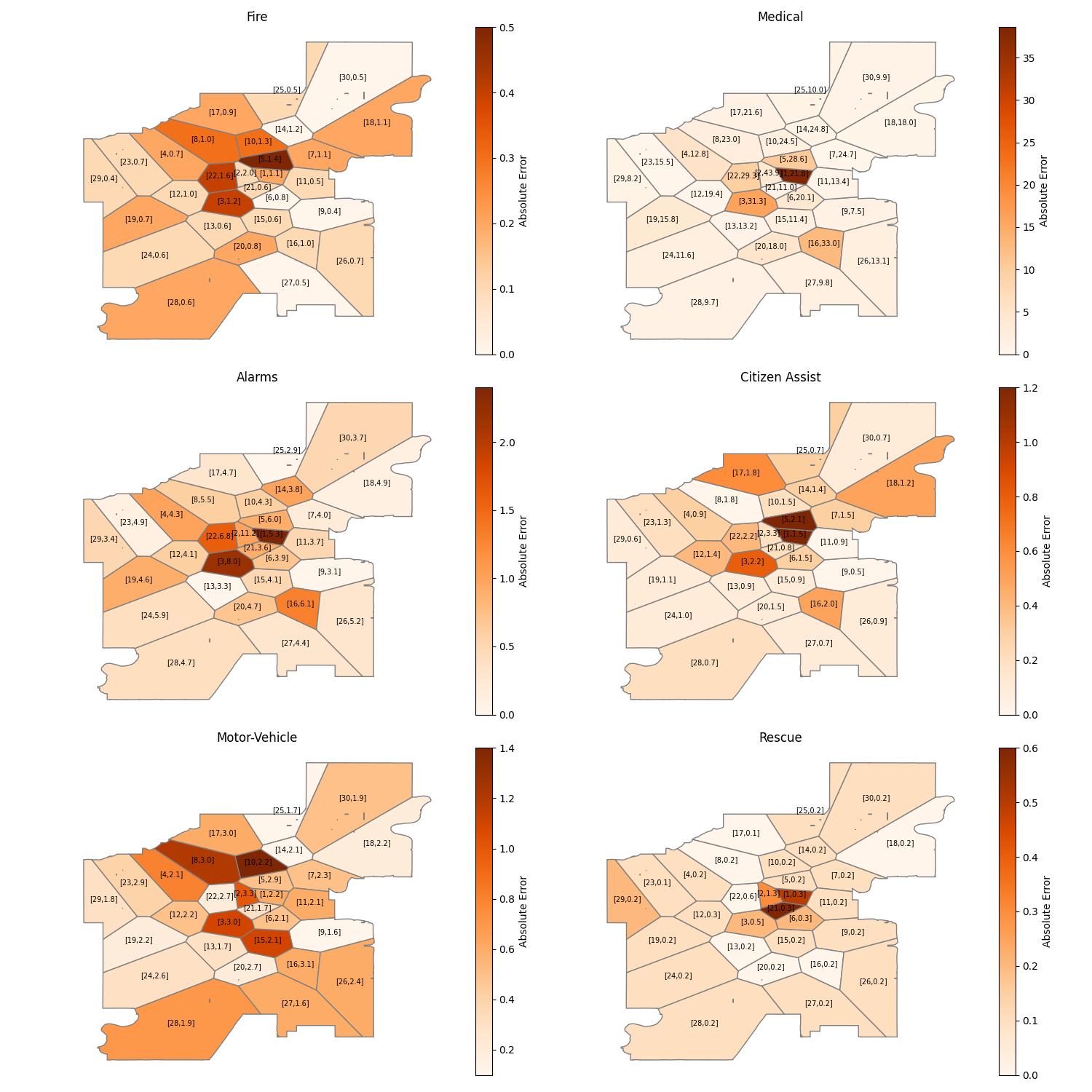}
\caption {\small {Fire-stations with weekly prediction error for different event categories. The annotation vector on the polygon map represents a fire station number (first element), and the mean of the weekly predicted events (second element), respectively.}}
\label{fig_weekly_error_map}
\end{figure*}

\begin{figure*} [hbt!]
		\centering
 \includegraphics[width=1.0\textwidth,keepaspectratio]{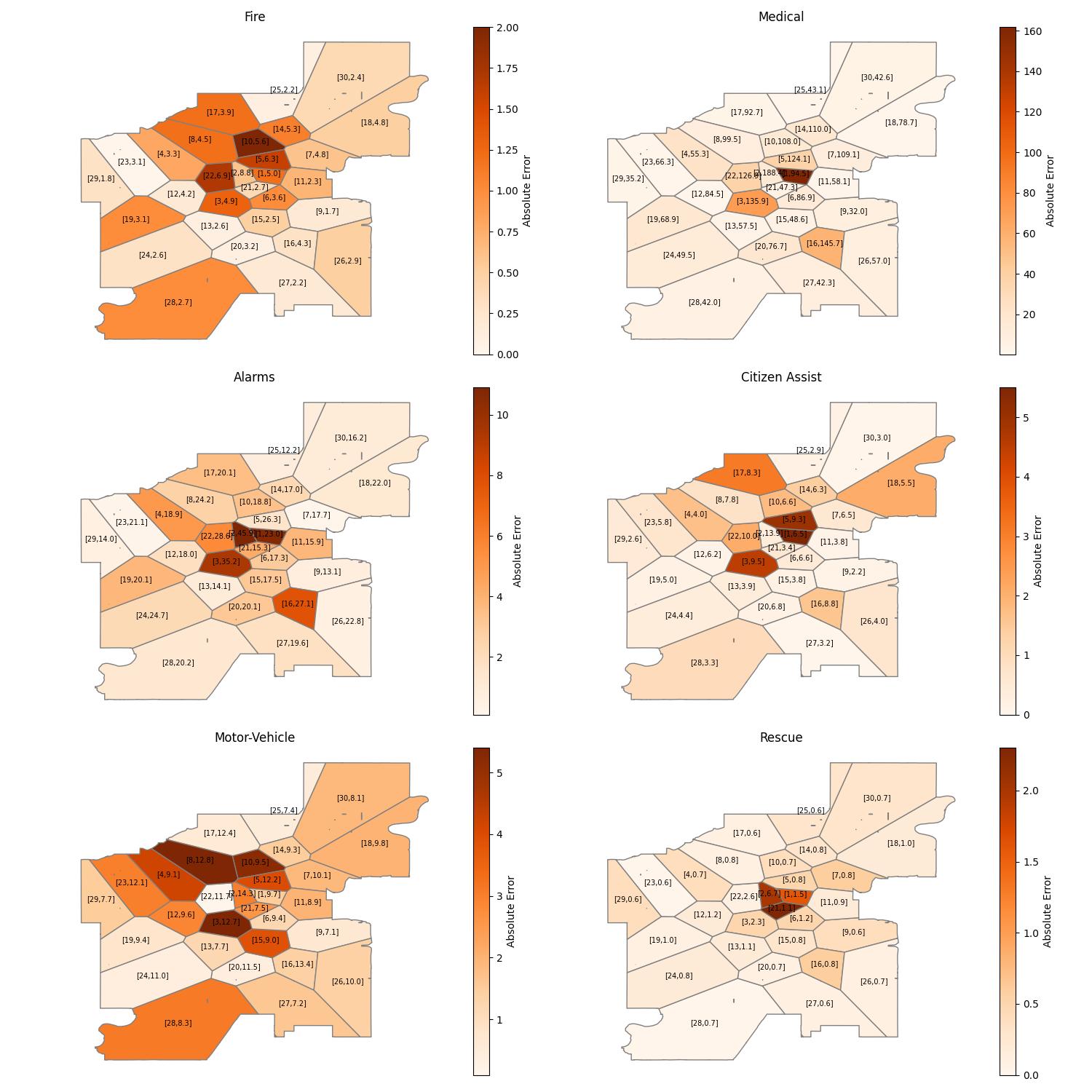}
\caption {\small {Fire stations with monthly prediction error for different event categories. The annotation vector on the polygon map represents a fire station number (first element), and the mean of the monthly predicted events (second element), respectively.}}
\label{fig_monthly_error_map}
\end{figure*}

\subsection{Model Evaluation \&  Prediction Error Analysis} 

We evaluated the performance of the predictive models of different event categories for each fire station. We analyzed the prediction error of each fire station for the out-of-sample test data.  There are many measures (e,g, Mean Absolute Error (MAE), Root Mean Squared Error (RMSE), Mean Absolute Percentage Error (MAPE), Akaike Information Criterion (AIC) or Bayesian Information Criterion (BIC)) that can be used to assess the performance of the regression models. For our work, we decided to use absolute error or MAE and RMSE metrics because MAPE is be applicable as there are many event types with zero actual event incidents. Similarly, AIC and BIC are more appropriate for comparing models with the same dataset, however, we have different event data for different event types and durations.

 We computed the absolute error (residual) values and used them to determine how well the model performed. An absolute error of each station is estimated and compared to the model performance of each station. We also evaluated the overall performance of the models using common regression metrics such as mean absolute error (MAE), root mean square error (RMSE), and absolute error. An absolute error of the model in a fire station is an absolute difference between the mean predicted value and the mean actual value.  Figure~\ref{fig_weekly_error_map}, and Figure~\ref{fig_monthly_error_map} show the plots of the weekly and monthly prediction errors for each event type in fire-station service regions. These results show that the performance of the models varies station by station because each station has different characteristics.

The error range of the model for weekly prediction of fire events ranges from 0 to 0.5 absolute error. This shows that the fire event prediction model performed well as we have a very small error range for most stations.  However, the model has more errors with stations 5, 19, 22 and 28. This shows that the model not performing well for those stations because these stations can have some special characteristics that influence fire event occurrence. 

Similarly, the mean of the absolute error of weekly and monthly prediction models of medical events by fire station are 0 to 36 and  0 to 170, respectively. Except for three fire stations 1, 3, and 16, the errors of other stations are smaller. This model has a very low error of approximately $0$ for stations 7, 9, 11, 12, 13, 17, and 23, and it shows that it performed very well for these stations. Station 3 has an interesting error, which shows that this station can have some different characteristics pertaining to its location.

Table \ref{tab:mae_vs_full_duration} presents the weekly and monthly predictive model performance by fire station for six different event types using MAE and RMSE metrics. As noted above, the prediction error of some of the stations is a relatively high because each of these stations has certain distinctive characteristics pertaining to their locations. The medical event prediction model does not perform well in stations 1 and 5 as these two stations seem to have specific features. There is a relatively high error with stations 1, 2, and 3 for alarm events because these stations are likely due to being in the central part of the city with more population density. Like the medical events, there is a relatively high error in stations 1, 2, and 5 for citizen-assist events. There is an interesting and different result with motor vehicle incidents because this type of event is more likely to happen in areas with more traffic density, worse road conditions, or bad weather. Fire station 21 has a significantly large error for rescue events as its location is near the river. Similarly, almost the same number of rescue events occurred in station 1. This clearly shows that there should be some association between station 1 and station 21, particularly for the rescue events. We can improve the prediction performance for these stations by investigating the factors specific to these stations.

\begin{table*}[]
\caption{Comparing Model Performance by Fire Station with MAE and RMSE Metrics.}
\resizebox{.95\textwidth}{!}{
\label{tab:mae_vs_full_duration}
\begin{tabular}{|c|c|cccccccccccc|}
\hline
\cellcolor[HTML]{FFFFFF} &
  \cellcolor[HTML]{FFFFFF} &
  \multicolumn{12}{c|}{Event Types} \\ \cline{3-14} 
\rowcolor[HTML]{FFFFFF} 
\cellcolor[HTML]{FFFFFF} &
  \cellcolor[HTML]{FFFFFF} &
  \multicolumn{2}{c|}{\cellcolor[HTML]{FFFFFF}Fire} &
  \multicolumn{2}{c|}{\cellcolor[HTML]{FFFFFF}Medical} &
  \multicolumn{2}{c|}{\cellcolor[HTML]{FFFFFF}Alarms} &
  \multicolumn{2}{c|}{\cellcolor[HTML]{FFFFFF}Citizen-Assist} &
  \multicolumn{2}{c|}{\cellcolor[HTML]{FFFFFF}Motor-Vehicle} &
  \multicolumn{2}{c|}{\cellcolor[HTML]{FFFFFF}Rescue} \\ \cline{3-14}  

\rowcolor[HTML]{FFFFFF} 
\multirow{-3}{*}{\cellcolor[HTML]{FFFFFF}Station   Number} &
  \multirow{-3}{*}{\cellcolor[HTML]{FFFFFF}Model} &
  \multicolumn{1}{c|}{\cellcolor[HTML]{FFFFFF}MAE} &
  \multicolumn{1}{c|}{\cellcolor[HTML]{FFFFFF}RMSE} &
  \multicolumn{1}{c|}{\cellcolor[HTML]{FFFFFF}MAE} &
  \multicolumn{1}{c|}{\cellcolor[HTML]{FFFFFF}RMSE} &
  \multicolumn{1}{c|}{\cellcolor[HTML]{FFFFFF}MAE} &
  \multicolumn{1}{c|}{\cellcolor[HTML]{FFFFFF}RMSE} &
  \multicolumn{1}{c|}{\cellcolor[HTML]{FFFFFF}MAE} &
  \multicolumn{1}{c|}{\cellcolor[HTML]{FFFFFF}RMSE} &
  \multicolumn{1}{c|}{\cellcolor[HTML]{FFFFFF}MAE} &
  \multicolumn{1}{c|}{\cellcolor[HTML]{FFFFFF}RMSE} &
  \multicolumn{1}{c|}{\cellcolor[HTML]{FFFFFF}MAE} &
  RMSE \\ \hline
\rowcolor[HTML]{FFFFFF} 
\cellcolor[HTML]{FFFFFF} &
  Weekly &
  \multicolumn{1}{c|}{\cellcolor[HTML]{FFFFFF}0.90} &
  \multicolumn{1}{c|}{\cellcolor[HTML]{FFFFFF}1.19} &
  \multicolumn{1}{c|}{\cellcolor[HTML]{FFFFFF}37.57} &
  \multicolumn{1}{c|}{\cellcolor[HTML]{FFFFFF}47.12} &
  \multicolumn{1}{c|}{\cellcolor[HTML]{FFFFFF}2.98} &
  \multicolumn{1}{c|}{\cellcolor[HTML]{FFFFFF}4.74} &
  \multicolumn{1}{c|}{\cellcolor[HTML]{FFFFFF}1.94} &
  \multicolumn{1}{c|}{\cellcolor[HTML]{FFFFFF}2.65} &
  \multicolumn{1}{c|}{\cellcolor[HTML]{FFFFFF}1.34} &
  \multicolumn{1}{c|}{\cellcolor[HTML]{FFFFFF}1.60} &
  \multicolumn{1}{c|}{\cellcolor[HTML]{FFFFFF}0.73} &
  1.00 \\
\hline
\end{tabular}
}
\end{table*}

\section{Impact of COVID-19 Pandemic on Event Predictions}
\label{sec-covidimpact}
 The occurrence patterns of fire and other emergency events changed significantly during the COVID-19 pandemic. The impact of the pandemic on city dynamics varied according to conditions such as lockdowns and local restrictions, population density, and socioeconomic activities, including business and industrial operations. Many business and commercial activities in the city were reduced, and some of these activities shifted to residential areas due to lockdowns and other restrictions. This situation led to the occurrence of more events in residential areas or suburbs of the city. In addition, event occurrence patterns may have fluctuated over the time of the pandemic as local restrictions were implemented and removed. These situations can impact the performance of the emergency event predictive models. In this section, we present the results of how the COVID-19 pandemic situation impacted EFRS event predictions with our model. 
 
 We conducted several experiments for each event category with the same demographic and socioeconomic features of fire station regions. In these experiments, we trained and evaluated the models before and after the COVID-19 pandemic periods of weekly and monthly events data from 2017-05-01 to 2019-12-30, and 2020-01-01 to 2022-08-01, respectively. As before, we randomly split the data into train and test samples with 70\% of data for training and 30\% of data for testing. We trained the models and validated them with out-sample test event data. The performance of both weekly and monthly predictive models of each event type was evaluated with MAE and RMSE for before and after the pandemic periods.

\begin{figure*} [hbt!]
		\centering
 \includegraphics[width=1.0\textwidth,keepaspectratio]{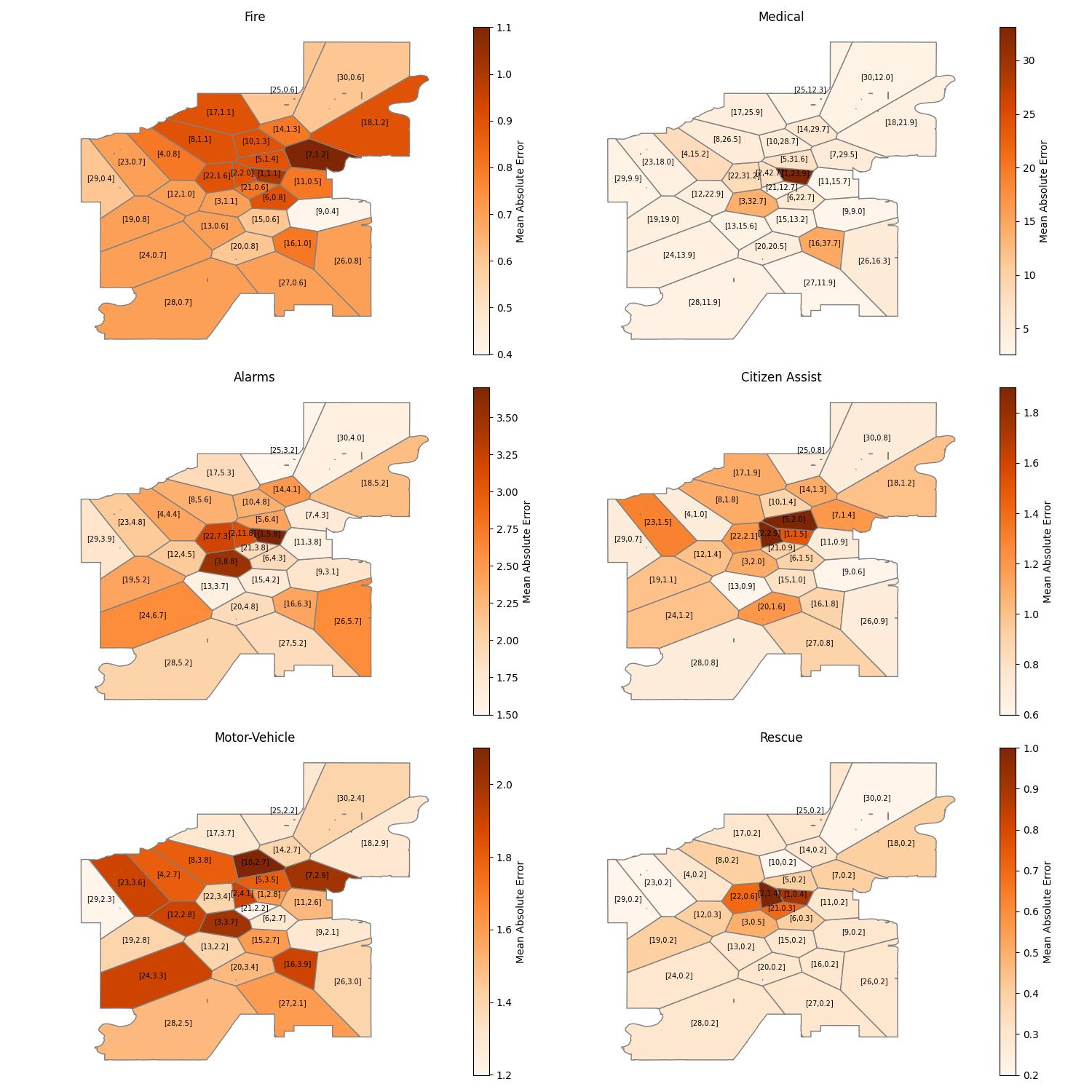}
\caption {\small {Weekly prediction of events by fire-stations before COVID-19 with mean absolute error (MAE) for different event types. The annotation vector on the polygon map represents a fire station number (first element), and the mean of the weekly predicted events (second element), respectively.}}
\label{fig_weekly_error_before_map}
\end{figure*}

\begin{figure*} [hbt!]
		\centering
 \includegraphics[width=1.0\textwidth,keepaspectratio]{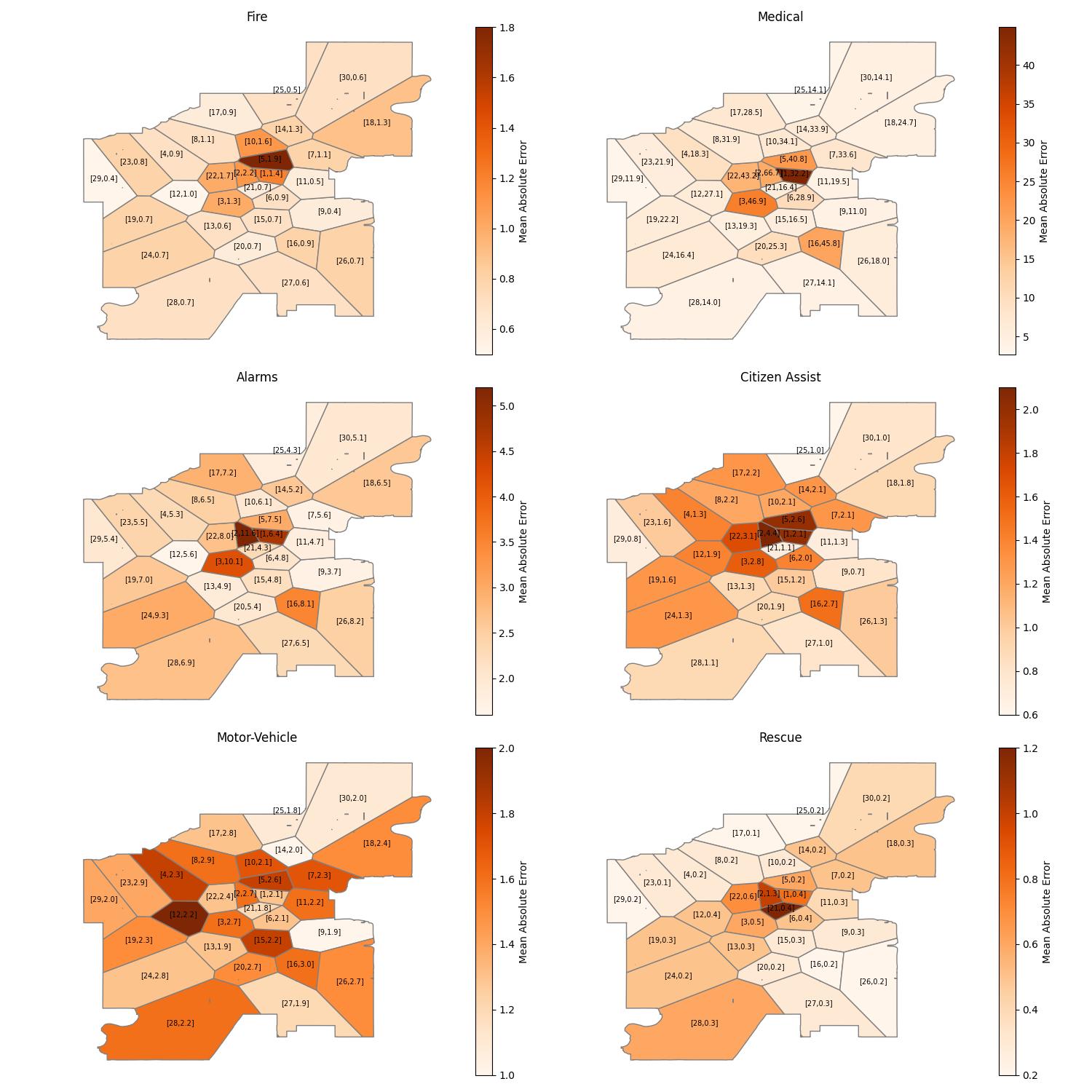}
\caption {\small {Weekly prediction of events by fire-station after COVID-19 with mean absolute error (MAE) for the different event types. The annotation vector on the polygon map represents a fire station number (first element), and the mean of the weekly predicted events (second element), respectively.}}
\label{fig_weekly_error_after_map}
\end{figure*}

Figure~\ref{fig_weekly_error_before_map} shows the heap-map plot of prediction results with mean absolute error using before the COVID-19 pandemic weekly event predictive model. Similarly, Figure~\ref{fig_weekly_error_after_map} depicts the weekly prediction of events by fire station after the COVID-19 pandemic with mean absolute error (MAE) for the different event types. The annotation vector on the polygon map represents a fire station number (first element), and the mean of the weekly predicted events (second element), respectively. Similarly, Table \ref{tab:overall_performance_before-after} presents the comparison of prediction error metric values for assessing the overall performance of the predictive models before and after the COVID-19 pandemic. Table ~\ref{tab:rmsebeforeafter} shows the model performance comparison by fire station before and after the COVID-19 pandemic using RMSE. 

When we trained the models using before COVID-19 and after COVID-19 events data, we obtained slightly different models (predictors' coefficients) for different event types. Thus we have slightly different models for making predictions before and after the pandemic period. However, the results show that there is a relatively high error rate with the models using the data after COVID-19. All of the results tell us that there were significant changes in the occurrence of events both in terms of locations and volumes during the COVID-19 pandemic. In addition, there are some stations such as stations 1, 2, 3, 5, 8, and 21  that have relatively high error after-COVID-19 models because some of these stations have special characteristics in terms of their locations and their closer proximity to public essential services (e.g., hospitals, shelters, and pharmacies) which can lead to the likelihood of occurrence of more emergency events. This indicates that there was a significant impact of the COVID-19 pandemic situation on the performance of the models for the occurrence of fire and other emergency events in the city. 

We can improve the predictions of the models by exploring other explanatory dynamic factors that impact the occurrence of the events during the pandemic. Some of the potential dynamic factors of the suburban areas can be identified by keeping track of changes in home occupancy, residential activities, commercial activities, and people commuting to the urban centers. In addition, network activity data could help in capturing the people's mobility in the city. We can collect these data by counting the mobile phones connected to to nearby cell towers of the cellular network. 

\begin{table}[]
\centering
\caption{\small{Comparison of prediction errors for assessing the overall performance of the predictive models before and after COVID-19}}
\label{tab:overall_performance_before-after}
\resizebox{.48\textwidth}{!}{
\begin{tabular}{|c|c|cc|cc|}
\hline
\rowcolor[HTML]{FFFFFF} 
\cellcolor[HTML]{FFFFFF} &
  \cellcolor[HTML]{FFFFFF} &
  \multicolumn{2}{c|}{\cellcolor[HTML]{FFFFFF}} &
  \multicolumn{2}{c|}{\cellcolor[HTML]{FFFFFF}} \\
\rowcolor[HTML]{FFFFFF} 
\cellcolor[HTML]{FFFFFF} &
  \cellcolor[HTML]{FFFFFF} &
  \multicolumn{2}{c|}{\multirow{-2}{*}{\cellcolor[HTML]{FFFFFF}MAE}} &
  \multicolumn{2}{c|}{\multirow{-2}{*}{\cellcolor[HTML]{FFFFFF}RMSE}} \\ \cline{3-6} 
\rowcolor[HTML]{FFFFFF} 
\multirow{-3}{*}{\cellcolor[HTML]{FFFFFF}\begin{tabular}[c]{@{}c@{}}Event \\ Types\end{tabular}} &
  \multirow{-3}{*}{\cellcolor[HTML]{FFFFFF}Model} &
  \multicolumn{1}{c|}{\cellcolor[HTML]{FFFFFF}\begin{tabular}[c]{@{}c@{}}Before \\ COVID-19\end{tabular}} &
  \begin{tabular}[c]{@{}c@{}}After \\ COVID-19\end{tabular} &
  \multicolumn{1}{c|}{\cellcolor[HTML]{FFFFFF}\begin{tabular}[c]{@{}c@{}}Before\\  COVID-19\end{tabular}} &
  \begin{tabular}[c]{@{}c@{}}After\\  COVID-19\end{tabular} \\ \hline
\rowcolor[HTML]{FFFFFF} 
\cellcolor[HTML]{FFFFFF} &
  Weekly &
  \multicolumn{1}{c|}{\cellcolor[HTML]{FFFFFF}0.77} &
  0.83 &
  \multicolumn{1}{c|}{\cellcolor[HTML]{FFFFFF}0.95} &
  1.10 \\ 
\multirow{-2}{*}{\cellcolor[HTML]{FFFFFF}Fire} &
  \cellcolor[HTML]{FFFFFF}Monthly &
  \multicolumn{1}{c|}{1.57} &
  \cellcolor[HTML]{FFFFFF}1.87 &
  \multicolumn{1}{c|}{\cellcolor[HTML]{FFFFFF}1.85} &
  \cellcolor[HTML]{FFFFFF}2.66 \\ \hline
\rowcolor[HTML]{FFFFFF} 
\cellcolor[HTML]{FFFFFF} &
  Weekly &
  \multicolumn{1}{c|}{\cellcolor[HTML]{FFFFFF}6.01} &
  10.23 &
  \multicolumn{1}{c|}{\cellcolor[HTML]{FFFFFF}7.07} &
  18.20 \\ 
\rowcolor[HTML]{FFFFFF} 
\multirow{-2}{*}{\cellcolor[HTML]{FFFFFF}Medical} &
  Monthly &
  \multicolumn{1}{c|}{\cellcolor[HTML]{FFFFFF}21.79} &
  37.45 &
  \multicolumn{1}{c|}{\cellcolor[HTML]{FFFFFF}24.44} &
  70.92 \\ \hline
\rowcolor[HTML]{FFFFFF} 
\cellcolor[HTML]{FFFFFF} &
  Weekly &
  \multicolumn{1}{c|}{\cellcolor[HTML]{FFFFFF}2.18} &
  2.55 &
  \multicolumn{1}{c|}{\cellcolor[HTML]{FFFFFF}2.70} &
  3.43 \\ 
\rowcolor[HTML]{FFFFFF} 
\multirow{-2}{*}{\cellcolor[HTML]{FFFFFF}Alarms} &
  Monthly &
  \multicolumn{1}{c|}{\cellcolor[HTML]{FFFFFF}5.77} &
  6.54 &
  \multicolumn{1}{c|}{\cellcolor[HTML]{FFFFFF}7.09} &
  8.53 \\ \hline
\rowcolor[HTML]{FFFFFF} 
\cellcolor[HTML]{FFFFFF} &
  Weekly &
  \multicolumn{1}{c|}{\cellcolor[HTML]{FFFFFF}1.02} &
  1.23 &
  \multicolumn{1}{c|}{\cellcolor[HTML]{FFFFFF}1.29} &
  1.71 \\ 
\rowcolor[HTML]{FFFFFF} 
\multirow{-2}{*}{\cellcolor[HTML]{FFFFFF}\begin{tabular}[c]{@{}c@{}}Citizen\\  Assist\end{tabular}} &
  Monthly &
  \multicolumn{1}{c|}{\cellcolor[HTML]{FFFFFF}2.77} &
  3.62 &
  \multicolumn{1}{c|}{\cellcolor[HTML]{FFFFFF}3.18} &
  5.03 \\ \hline
\rowcolor[HTML]{FFFFFF} 
\cellcolor[HTML]{FFFFFF} &
  Weekly &
  \multicolumn{1}{c|}{\cellcolor[HTML]{FFFFFF}1.59} &
  1.44 &
  \multicolumn{1}{c|}{\cellcolor[HTML]{FFFFFF}2.00} &
  1.83 \\ 
\rowcolor[HTML]{FFFFFF} 
\multirow{-2}{*}{\cellcolor[HTML]{FFFFFF}\begin{tabular}[c]{@{}c@{}}Motor \\ Vehicle\end{tabular}} &
  Monthly &
  \multicolumn{1}{c|}{\cellcolor[HTML]{FFFFFF}4.33} &
  3.66 &
  \multicolumn{1}{c|}{\cellcolor[HTML]{FFFFFF}5.19} &
  4.66 \\ \hline
\rowcolor[HTML]{FFFFFF} 
\cellcolor[HTML]{FFFFFF} &
  Weekly &
  \multicolumn{1}{c|}{\cellcolor[HTML]{FFFFFF}0.39} &
  0.45 &
  \multicolumn{1}{c|}{\cellcolor[HTML]{FFFFFF}0.44} &
  0.67 \\ 
\rowcolor[HTML]{FFFFFF} 
\multirow{-2}{*}{\cellcolor[HTML]{FFFFFF}Rescue} &
  Monthly &
  \multicolumn{1}{c|}{\cellcolor[HTML]{FFFFFF}0.93} &
  1.18 &
  \multicolumn{1}{c|}{\cellcolor[HTML]{FFFFFF}1.13} &
  1.98 \\ \hline
\end{tabular}
}
\end{table}

\begin{table*}[]
\centering
\caption{Comparing the Prediction Errors with RMSE Metric for Before and After COVID-19}
\resizebox{.95\textwidth}{!}{
\label{tab:rmsebeforeafter}
\begin{tabular}{|c|c|cccccccccccc|}
\hline
\cellcolor[HTML]{FFFFFF} &
  \cellcolor[HTML]{FFFFFF} &
  \multicolumn{12}{c|}{Event Types} \\ \cline{3-14} 
\rowcolor[HTML]{FFFFFF} 
\cellcolor[HTML]{FFFFFF} &
  \cellcolor[HTML]{FFFFFF} &
  \multicolumn{2}{c|}{\cellcolor[HTML]{FFFFFF}Fire} &
  \multicolumn{2}{c|}{\cellcolor[HTML]{FFFFFF}Medical} &
  \multicolumn{2}{c|}{\cellcolor[HTML]{FFFFFF}Alarms} &
  \multicolumn{2}{c|}{\cellcolor[HTML]{FFFFFF}Citizen-Assist} &
  \multicolumn{2}{c|}{\cellcolor[HTML]{FFFFFF}Motor-Vehicle} &
  \multicolumn{2}{c|}{\cellcolor[HTML]{FFFFFF}Rescue} \\ \cline{3-14} 
\rowcolor[HTML]{FFFFFF} 
\multirow{-3}{*}{\cellcolor[HTML]{FFFFFF}Station   Number} &
  \multirow{-3}{*}{\cellcolor[HTML]{FFFFFF}Model} &
  \multicolumn{1}{c|}{\cellcolor[HTML]{FFFFFF}Before} &
  \multicolumn{1}{c|}{\cellcolor[HTML]{FFFFFF}After} &
  \multicolumn{1}{c|}{\cellcolor[HTML]{FFFFFF}Before} &
  \multicolumn{1}{c|}{\cellcolor[HTML]{FFFFFF}After} &
  \multicolumn{1}{c|}{\cellcolor[HTML]{FFFFFF}Before} &
  \multicolumn{1}{c|}{\cellcolor[HTML]{FFFFFF}After} &
  \multicolumn{1}{c|}{\cellcolor[HTML]{FFFFFF}Before} &
  \multicolumn{1}{c|}{\cellcolor[HTML]{FFFFFF}After} &
  \multicolumn{1}{c|}{\cellcolor[HTML]{FFFFFF}Before} &
  \multicolumn{1}{c|}{\cellcolor[HTML]{FFFFFF}After} &
  \multicolumn{1}{c|}{\cellcolor[HTML]{FFFFFF}Before} &
  After \\
\hline
\end{tabular}
}
\end{table*}
%

\section{Key Insights \& Findings} \label{sec:insights}
In this section, we discuss how our findings answer the research questions Q1 to Q8. We obtained the following insights and findings from this study that answer to the research questions:
\begin{enumerate}

    \item {\bf Potential data sources:} We identified the potential data sources for fire and emergency event prediction, namely, Open Data Portal of the City, Open Street Map, and Fire Rescue Service Department from where we collected demographic, socioeconomic, point of interest (PoI) data, and EFRS event incidents data, respectively.
    
    \item {\bf Dataset development:} We developed a methodology to prepare the datasets at different spatio-temporal granularities (e.g., neighborhoods and fire-station regions; and daily, weekly, and monthly). We developed computational methods and built tools to collect data from different sources, mapped fire-station service areas, and then converted neighborhood characteristics to fire-stations.    
    
    \item {\bf Correlation analysis:} We identified the following findings from the correlation analysis of event types with neighborhood features
    \begin {itemize}
        \item Fire events: low-income, apartments/condos with 1-4 stories, and the number of hotels/motels are the top three features that are correlated with fire events. This means that the likelihood of fire incidents increases as the number of low-income populations, condos/apt, and hotels/motels increase. 
        \item Medical events: medical events are highly correlated with the level of income. The next correlated feature is more crowded places such as food-related facilities and hotels/motels. This indicates that the increasing low-income population, and food-related facilities (restaurants, bars, food courts, and hotels/motels) in a neighborhood can cause the chance of occurring more medical event incidents.
 
        \item Alarm events: apartments/condos with more than 5 stories, food-related facilities, hotels/motels, and low income are the highly correlated features with alarm events. This shows that the alarms are likely to happen in neighborhoods with more apartments or condominiums, food-related facilities, hotels/motels, and low-income populations (e.g., Downtown).

        \item Citizen assist events: low-income, food, and condos/apartments are highly correlated with citizen assist events. This shows the level of income,  more populated places, and food-related facilities of a neighborhood impact the likelihood of the occurring citizen assist event incidents. 

        \item Motor vehicle events: no. of traffic lights, food, income, and commercial are the highly correlated features with motor vehicle incidents. These events are likely to happen in areas with more traffic lights, food facilities, commercial places, and low-income populations.

        \item Rescue events: apartments/condos with 5 or more stories, food facilities, and hotels/motels are three highly correlated features with rescue events. This shows that rescue events are more likely to happen around more populated places such as apartments/condos with 5 or more stories, food facilities, and hotels/motels.
    \end{itemize}

\item {\bf Predictive model development using neighborhood features:} We developed the linear regression models for weekly and monthly event prediction by neighborhood and fire station using characteristics such as demographic, socioeconomic, and point of interest features. These are generic predictive models built with the identified important feature sets (predictors) for the events. We developed models for predicting the events in the following two spatial resolutions:
\begin{itemize}
    \item At the neighborhood level, two models were developed for predicting the likelihood of the occurrence of total the weekly or monthly emergency events with its given features (e.g., income, food-related facilities, apartments/condos, no. of traffic lights, length of terms of residency, etc.). These models performed well with an acceptable error rate.  However, the models can not make a good prediction for an individual event type at a neighborhood level because of their low occurrence rate. 

    \item At a fire station level, we developed different models for different event types with their corresponding identified importance features (predictors). For example, a model for a fire event was developed with low and medium-income, apartments/condos 1 to 4 stories, populations with no certificate or diploma or degree (no education), and less than one-year term length features (predictors). The coefficients of these features were estimated during the training of the models, which were then used for the predictions. Except for rescue events, our developed models performed accurate predictions at the fire-station level for all other event types including fire, medical, alarms, citizen-assist, and motor-vehicle incidents.
\end{itemize}
 
\item {\bf Event predictability analysis:} 
Based on our prediction error analysis, we summarized the findings on the predictability of events, and event types by neighborhood and fire station, respectively as follows: 
\begin{itemize}
\item For most neighborhoods, the events can be predicted with the given features such as household income, building types and occupancy, length terms residency, education, food-related facilities, retail, commercial, roads, and traffic information. A neighborhood with ID 1010 has a very high error and the models did not make have good prediction performance for it.

\item At a fire station level, each event type is predictable with the given features except for a few special fire stations. 
\begin{itemize}
    \item Fire events are predictable for all fire stations. However, station 5 and station 22 have a high prediction error rate.
    \item Medical events are relatively more predictable and the predictions are more accurate except for station 1. This station has an extremely high number of actual medical event incidents. 
    \item Alarm events are predictable for most of the fire stations. However, the fire stations 1, 3, 16, and 22 have high prediction error.
    \item Citizen-assist events are predictable for all fire stations except station 1. Also, medical and citizen-assist events have similar characteristics. 
    \item Motor vehicle rescue events are slightly less predictable with these features. The models could not make good predictions for fire stations 2, 3, 8, 10, 15, and 28.
    \item Although there is a small prediction error with rescue events, these events are relatively less predictable because these events rarely occur and mostly happen in fire stations 1 and 21 which have a high error rate. It also found that these two fire stations have some association, particularly for the rescue events.
\end{itemize}
\end{itemize}

\item {\bf Spatial analysis of event predictability:} The predictions with the given features for the bigger spatial regions such as fire station coverage areas are more accurate and predictable than smaller neighborhood areas. The predictions of the likelihood of individual event types occurrence at a neighborhood level are difficult because the majority of the event types occur rarely and also there is a very high degree of spatial variation. 

\item {\bf Temporal analysis of event predictability:} The predictions of the monthly events are more accurate and predictable than weekly events. This shows that the prediction of an individual event type for a shorter time interval (e.g., hourly, daily) would be difficult because the majority of the event types occur rarely and also there is a very high degree of temporal variations.

\item {\bf Impact of COVID-19 pandemic on event prediction:} During COVID-19, there is a significant impact of the COVID-19 pandemic on event prediction because there were significant changes in the pattern of occurrence of events both in terms of locations and frequency. The prediction performance of the models with after-COVID-19 data is slightly lower than before COVID-19. In particular, stations 1, 2, 3, 5, 8, and 21 have relatively higher errors with after-COVID-19 models. There are various factors that impact the occurrences of the events during the COVID-19 pandemic. There were huge changes in people's mobility and other activities which impacted the occurrence of the emergency events. Identifying these changing patterns, and incorporating them into the models can help to improve the performance of the predictive models during or after the pandemic.
\end{enumerate}

\section{Conclusions \& Future Work} \label{section-conclusion}
 In this paper, we develop statistical analysis and machine learning models to analyze and predict EFRS events for the city of Edmonton, Canada. We prepared datasets by collecting and joining diverse sources with different temporal periods at neighborhood and fire-station spatial levels that can be used for EFRS events prediction. We analyzed the characteristics of EFRS events and their occurrence pattern at different granularities. We also provided descriptive statistics and feature associations. We developed and built a number of predictive models for each EFRS event using their key feature set. We made predictions with the developed models and validated the results by comparing the predicted and actual event occurrences. We analyzed the impact of COVID-19 pandemic situation on EFRS events prediction and assessed the performance of the predictive models for the pandemic period by comparing its prediction error metric values with before the COVID-19. The result of our work can help EFRS in monthly and weekly decision-making, prevention, and mitigation strategies at city, neighborhood, and fire station levels. The methodologies developed in our work have wide applicability to other cities as the data sources and features used are fairly standard.  The accuracy of the models is strongly influenced by the net event occurrence rates as well as spatial and temporal extent.

In future work, our plan includes (i) investigation of weather, network activity, and traffic data for building the predictive models; (ii) building robust predictive models can aid in planning new fire stations in the areas of the city that have a high incident rate and/or require faster response time; (iii) development of multivariate time series forecasting model that can help us to identify the possible number of events happening in the future for a short interval of time (hour, day). A multivariate time series forecasting model with weather or any other dynamic features would be helpful in predicting different events. In addition, it can be used to detect anomalies using a confidence band for the predicted values and actual data points.

\section*{Acknowledgement}
This work was supported by the funding research grants from TELUS and NSERC Alliance Grant \#576812.

\bibliographystyle{IEEEtranN} 

\bibliography{references}

\end{document}